\documentclass[a4paper,fleqn]{cas-sc}

\usepackage[authoryear]{natbib}
\usepackage{amssymb}
\usepackage{amsmath}
\usepackage{booktabs}
\usepackage{subcaption}
\usepackage{tikz}

\usepackage{multirow, graphicx}
\usepackage{hyperref}

\begin{document}
\let\WriteBookmarks\relax
\let\printorcid\relax

\shorttitle{}
\shortauthors{K. Z. Li et~al}

\title [mode = title]{Reliability-aware short-term roll prediction for unmanned surface vehicles via multi-task learning and adaptive centralization}

\author[label1]{Kaizhen Li} 
\fnmark[1]
\credit{Conceptualization, Methodology, Validation, Visualization, 
Writing - original draft}

\author[label2]{Xi Zhou} 
\fnmark[1]
\credit{Conceptualization, Methodology, Supervision, Formal analysis, Writing - review \& editing}

\author[label3]{Zihao Wang} 
\credit{Formal analysis, Funding acquisition}

\author[label1]{Dan Zhang} 
\credit{Formal analysis, Funding acquisition}

\author[label3]{Jianjian Liu} 
\credit{Formal analysis, Funding acquisition}

\author[label1]{Xiaowei Li}
\cormark[1]
\ead{xwli@staff.shu.edu.cn}
\credit{Methodology, Supervision, Funding acquisition}

\cortext[cor1]{Corresponding author}
\fntext[fn1]{These authors contributed equally to this work and should be considered co-first authors.}

\affiliation[label1]{organization={School of Mechatronic Engineering and Automation, Shanghai University},
            city={Shanghai},
            postcode={200444}, 
            country={China}}
\affiliation[label2]{organization={Shanghai Artificial Intelligence Laboratory},
            city={Shanghai},
            postcode={200032}, 
            country={China}}
\affiliation[label3]{organization={Institute of Artificial Intelligence, Shanghai University},
            city={Shanghai},
            postcode={200444}, 
            country={China}}

\begin{abstract}
Reliable roll prediction of unmanned surface vehicles (USVs) is essential for ensuring navigational safety and enhancing autonomous decision-making. While existing studies primarily focus on improving prediction accuracy, the quantification of prediction reliability remains insufficiently addressed. To bridge this gap, this paper proposes a reliability-aware prediction paradigm that integrates confidence assessment into the predictive pipeline. The architecture utilizes a multi-task learning structure where a shared feature extraction backbone feeds into dual heads: a regression head for precise roll prediction and a quantification head for confidence scoring. This configuration provides accurate prediction and corresponding confidence for risk-sensitive downstream tasks. In addition, an adaptive centralization strategy tailored for short-term real-time roll prediction is introduced to improve model generalization under varying operational conditions. Experiments conducted on a real-sea dataset demonstrate that the proposed method effectively quantifies the reliability of prediction results and maintains superior generalization under varying conditions, offering significant potential for practical engineering applications.
\end{abstract}



\maketitle

\section{Introduction}
\label{Introduction}
Small-displacement unmanned surface vehicles (USVs) are inherently susceptible to wave-induced perturbations, which often trigger high-amplitude and short-period roll motions. 
These intense oscillations significantly compromise the stability of onboard weapon systems \citep{zhang2024hybrid} and the safety of UAV launch and recovery missions \citep{tian2024uav}. To mitigate these risks, there is an urgent need for multi-step motion predictions embedded with explicit uncertainty quantification. Such probabilistic predictive states are essential for achieving risk-aware autonomous decision-making \citep{jensen2007efficient}-allowing onboard weapons to select high-fidelity, high-confidence firing windows and enabling UAVs to identify reliable contact timing during launch and recovery.

Prediction approaches for ship roll motion can be broadly classified into mechanism-based method and data-driven method. Mechanism-based methods as referred to here include formulation-based approaches derived from seakeeping theory, as well as numerical solvers such as computational fluid dynamics (CFD)\citep{liu2018predictions} and nonlinear potential flow models\citep{alexandersson2021prediction}. However, the high computational cost of these solvers and the difficulty in precisely identifying hydrodynamic parameters often limit their real-time applicability. In contrast, data-driven approaches, particularly neural networks, leverage robust nonlinear modeling to capture intricate patterns from historical data without requiring explicit physical parameters. Diez \citep{diez2024data} investigated data-driven ship attitude prediction methods using feedforward neural networks, recurrent neural networks, and dynamic mode decomposition techniques. Hou \citep{hou2024short} proposed a convolutional neural network(CNN) based approach for short-term prediction of ship wave-induced roll motion. Zhang and Zhou \citep{zhang2023data}, considering the coupling relationships among wave height, pitch, and roll, developed a CNN-LSTM network to extract features and improve roll prediction accuracy. Li\citep{li2024short} employed an encoder-decoder architecture based on bidirectional long short-term memory networks (Bi-LSTM) combined with a teacher-forcing mechanism to achieve short-term ship roll motion prediction. 
Recently Transformer architectures have advanced ship motion forecasting, such as Informer for long-sequence acceleration \citep{zhang2025multivariate}and multi-scale CNN-Transformers for multi-frequency local-global feature fusion\citep{SHI2025115787}--these methods typically focus on deterministic point predictions.
Relying solely on deterministic predictions without assessing their uncertainty can be hazardous for autonomous operations. Therefore, confidence assessment of roll motion forecasts is needed, which can provide risk-aware references for the autonomous decision-making task.

To address the inherent limitations of deterministic models, extensive research has focused on neural-network-based uncertainty quantification techniques to generate prediction intervals for regression tasks\citep{khosravi2010lower}. Yeo \citep{yeo2018rnn} directly predicted the probability density function of future time series, providing an effective tool for forecasting time series with latent probabilistic distributions. Guo et al. \citep{guo2022probabilistic} employed deep learning models in conjunction with dropout techniques to perform probabilistic prediction of wave-induced heave motion of a semi-submersible floating platform. Furthermore, Huang \citep{huang2026deterministic} hypothesized that future roll motion states conform to Gaussian or Student’s t-distributions and employed long short-term memory (LSTM) supervised by Maximum Likelihood Estimation (MLE) loss, thereby enabling the generation of confidence intervals for predictions. 
Despite these advancements, existing methodologies primarily focus on constructing prediction intervals to describe the uncertainty range of future states, but lack the capability to assess the confidence of each predicted point across multiple forecasting steps. To bridge this gap, we propose a reliability-aware prediction paradigm that integrates step-wise confidence assessment into the predictive pipeline.

Moreover, the operating conditions and sea states encountered by USVs evolve continuously during navigation, inevitably leading to pronounced data distribution drift. This phenomenon, where the underlying statistical properties of motion data, such as mean and variance, shift over time-significantly compromises the generalization of predictive models. To mitigate the impact of such distribution shifts, existing methodologies have generally followed two distinct technical trajectories: online model adaptation via real-time parameter updates and data-centric regularization through statistical normalization. For online model adaptation, Yin\citep{yin2017real} proposed a ship roll prediction scheme based on an adaptive sliding window to characterize the time-evolving dynamics of roll motion. Similarly, Yu\citep{yu2024online} developed an adaptive weighted ensemble learning framework where the model weights are updated in response to varying conditions to maintain high prediction accuracy. Han\citep{han2024deep} proposed an online learning scheme for recurrent neural networks within a sliding-window framework, enabling real-time adaptation and multi-step prediction. Wang et al. \citep{WANG2025} proposed an incremental learning approach to achieve online parameter updates, which exhibited outstanding performance under rapidly shifting conditions in real sea datasets. 
For data-centric regularization, Kim et al. \citep{kim2021reversible} proposed Reversible Instance Normalization (RevIN), which counters distribution drift by normalizing inputs and symmetrically restoring the original distribution at the output stage, thereby maintaining predictive stability. Fan et al. \citep{fan2023dish} proposed Dish-ts and Liu et al. \citep{liu2023adaptive} proposed San, both methods further expand this approach by employing learnable scaling factors and fine-grained data transformations to mitigate both intra and inter-series distribution shifts.

However, the roll dynamics of USVs are highly nonlinear and time-varying, characterized by short oscillation periods \citep{naaijen2009real}. Therefore, conventional online adaptation strategies and statistical normalization techniques typically rely on sufficient historical observations to update model parameters or obtain robust statistical estimates, making them less effective for small-scale USVs exposed to rapidly changing short-term disturbances.

To address the above two challenges---namely, (i) the lack of reliability quantification for each specific roll forecast and (ii) the degradation caused by distribution drift under changing operating conditions---this paper proposes a reliability-aware prediction paradigm with an adaptive centralization strategy.
As illustrated in Fig.~\ref{fig:overall_structure}, the proposed paradigm adopts a multi-task architecture in which a shared feature extraction backbone feeds into two heads: (1) a regression head for accurate roll prediction and (2) a confidence assessment (quantification) head for confidence scoring. Concretely, the confidence assessment head outputs a probability distribution over discretized roll-angle intervals, which enables step-wise confidence estimation and uncertainty characterization. To tightly couple prediction and confidence, a consistency loss is further introduced to align the regression output with the maximum-confidence interval inferred by the confidence assessment head. To address the degradation in model generalization caused by distribution drift under time-varying operating conditions and sea states, we propose an adaptive centralization method. Specifically, the input sequence is first de-mean centralized to suppress non-stationary offsets, and a trend prediction module is then employed to estimate and compensate for mean shifts between the training and online data, as well as between the input and output sequences. Extensive experiments on real-sea roll datasets collected under multiple operating conditions validate that the proposed method improves robustness and generalization under distribution drift.

\begin{figure}[pos=t]
\centering
    \includegraphics[width=0.9\linewidth]{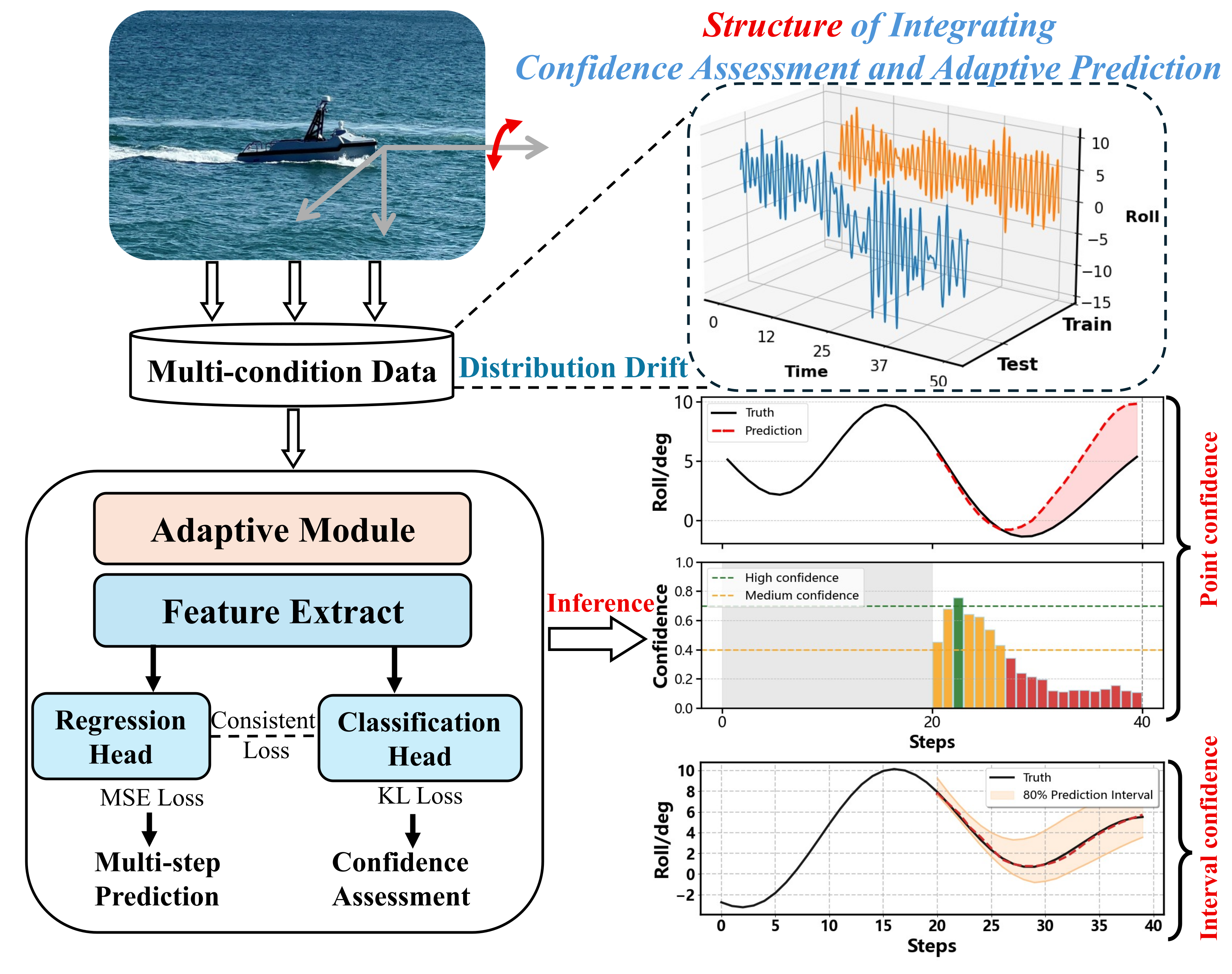}
    \caption{Overall structure}
    \label{fig:overall structure}
\end{figure}

The main contributions of this paper can be summarized as follows:
\begin{enumerate}
\item We propose a reliability-aware roll prediction paradigm that jointly outputs accurate multi-step roll prediction and confidence estimates for risk-sensitive downstream tasks.
\item We design a confidence assessment head and a regression--confidence consistency learning strategy, enabling calibrated confidence scoring and the construction of high-quality prediction intervals.
\item We propose an adaptive centralization strategy to mitigate distribution drift, and validate its generalization capability on real-sea roll datasets collected under multiple operating conditions.
\end{enumerate}

\section{Methodology}
\subsection{Adaptive Centralization}
\label{section: Adaptive Centralization}
Under complex sea conditions, time-varying operating conditions and sea states induce distribution drift in the roll motion of USVs, manifested as continuous shifts in the mean and scale. This drift not only occurs across different voyages (train--test/online shift) but can also appear within a single prediction window (input--output shift), which makes short-horizon forecasting particularly challenging.

A straightforward data-regularization choice is Z-Score standardization:
\begin{equation}
z_t = \frac{x_t - \mu}{\sigma},
\label{eq:z-score}
\end{equation}
where $x_t$ denotes the input sequence, $\mu$ and $\sigma$ are the mean and standard deviation, and $z_t$ is the normalized sequence. However, Z-Score is not always suitable for real-time USVs prediction with short input horizons: (i) accurate $\mu$ and $\sigma$ typically require long histories, whereas online prediction relies on short-window statistics; in stable segments, $\sigma$ may approach zero and excessively amplify the normalized values; (ii) statistics estimated offline from training data may not match online distributions; and (iii) it implicitly assumes consistent statistics between input and future output sequences.

\begin{figure}[pos=htbp]
    \centering
    \includegraphics[width=0.7\linewidth]{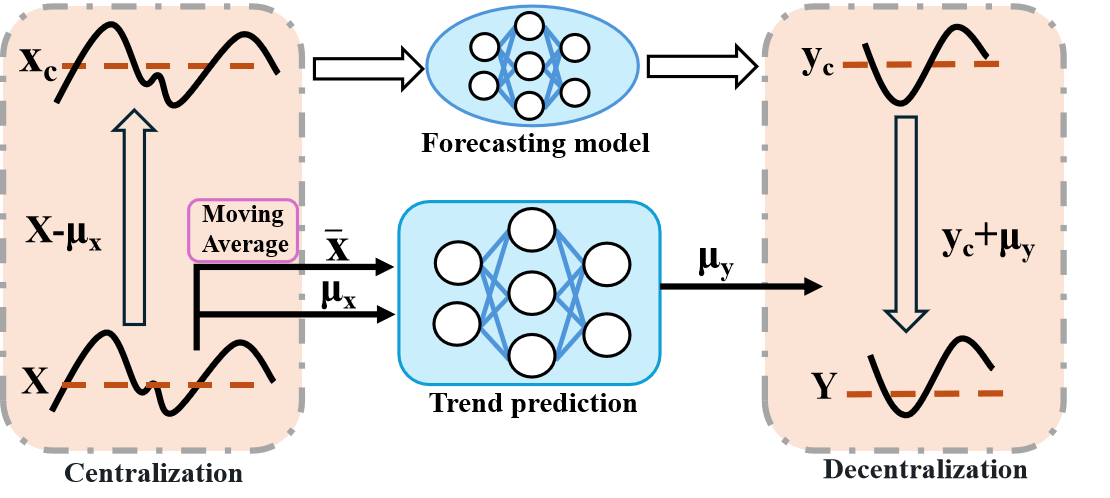}
    \caption{Adaptive centralization}
    \label{fig:adaptive_center}
\end{figure}

Motivated by adaptive regularization, we propose an adaptive centralization module that reduces the sensitivity to such drift while preserving the underlying dynamics. As illustrated in Fig.~\ref{fig:adaptive_center}, the module consists of two complementary components: (1) mean-based centralization to remove non-stationary offsets in the input sequence, and (2) a trend prediction module to estimate the mean shift between the input and output sequences, enabling adaptive centralization.

Given an input sequence $X$, we first compute its sample mean $\mu_x$ and obtain the centralized input $x_c$:
\begin{equation}
\begin{aligned}
\mu_x &= \frac{1}{T}\sum_{t=1}^{T} X_t,\\
x_{c} &= X - \mu_x.
\end{aligned}
\end{equation}
The centralized sequence $x_c$ is then fed into the forecasting model in Section~\ref{Prediction Paradigm} to predict the residual dynamics:
\begin{equation}
y_{c} = \mathrm{Forecasting\ Model}(x_{c}).
\end{equation}

To compensate for time-varying mean drift, we introduce a trend prediction module. Since short-term sequences are limited in length, a moving average filter is employed to extract a stable local trend representation:
\begin{equation}
\bar{x}_t = \frac{1}{k}
\sum_{i=t-\left\lfloor k/2 \right\rfloor}^{t+\left\lfloor k/2 \right\rfloor}
x_i,
\label{eq:Moving Average}
\end{equation}
where $k$ is the window size and boundary positions are padded by replicating the first and last values.

Considering the periodic nature of roll motion, we predict the future mean via a residual formulation relative to $\mu_x$:
\begin{equation}
\begin{aligned}
\delta &= \mathrm{MLP}(\bar{x} - \mu_x), \\
\mu_y &= \delta + \mu_x,
\end{aligned}
\label{eq:trend prediction}
\end{equation}
where $\mu_y$ denotes the predicted mean of the output sequence.

Finally, the de-centralized roll prediction is obtained as
\begin{equation}
Y = y_{c} + \mu_{y}.
\end{equation}

\subsection{Reliability-aware Prediction Paradigm}
\label{Prediction Paradigm}
\subsubsection{Feature Extraction Module}
To fully exploit the temporal dependencies of USVs roll motion, the encoder-decoder architecture based bidirectional long short-term memory network is introduced as a feature extraction module, providing informative representations for subsequent prediction and confidence estimation \citep{li2024short}.

\begin{figure}[pos=t]
    \centering
    \includegraphics[width=0.8\linewidth]{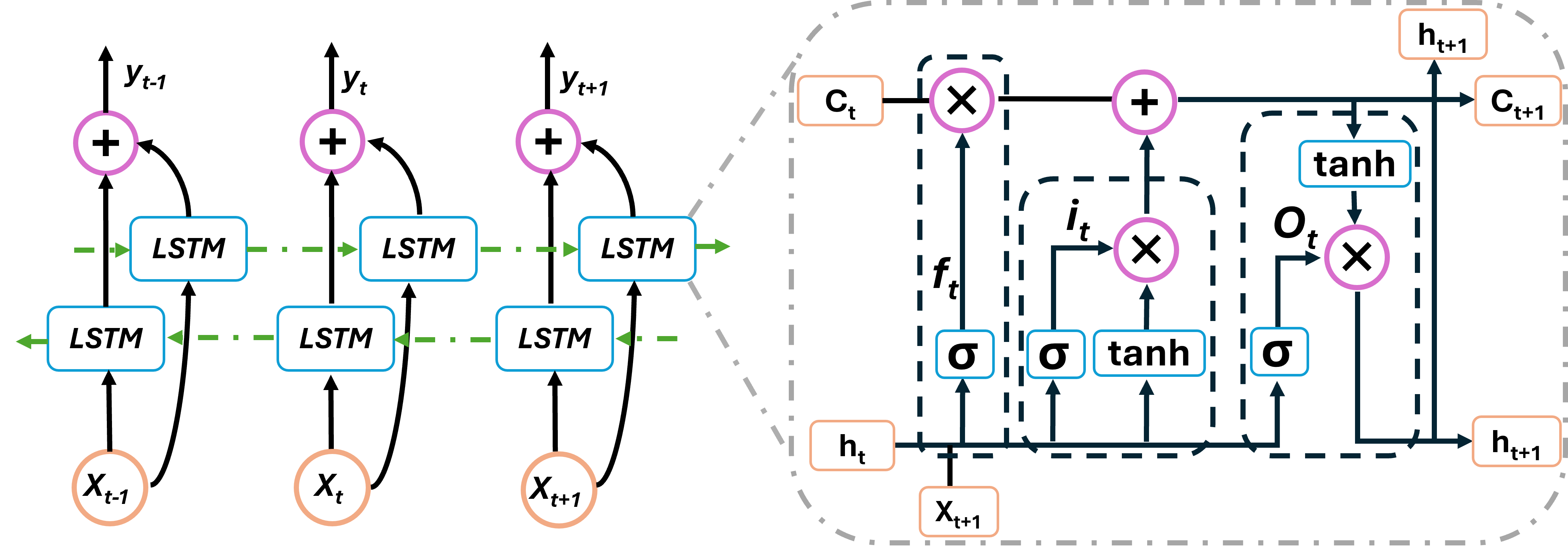}
    \caption{Structure of Bi-LSTM}
    \label{fig:Bi-LSTM}
\end{figure}

The bidirectional long short-term memory network Bi-LSTM \citep{graves2005framewise} is an extension of the traditional long short-term memory network \citep{hochreiter1997long, graves2012long} for bidirectional sequence modeling. LSTM introduces gating mechanisms to effectively mitigate the vanishing and exploding gradient problems commonly encountered in traditional recurrent neural networks (RNNs). Bi-LSTM further enhances the model’s ability to learn global sequence information by adding a backward LSTM branch to the original LSTM unidirectional information flow, enabling the model to leverage context information from both forward and backward temporal dimensions. The structure of the Bi-LSTM is illustrated in Fig.~\ref{fig:Bi-LSTM} and the mathematical formulation of this computation is given in Equation~\ref{eq:lstm}.

\begin{equation}
\begin{aligned}
f_t &= \sigma(W^{f,h}h_{t} + W^{f,x}x_{t+1} + b_f) \\
i_t &= \sigma(W^{i,h}h_{t} + W^{i,x}x_{t+1} + b_i) \\
o_t &= \sigma(W^{o,h}h_{t} + W^{o,x}x_{t+1} + b_o) \\
\tilde{c}_t &= \tanh(W^{c,h}h_{t} + W^{c,x}x_{t+1} + b_c) \\
c_{t+1} &= f_t \odot c_{t} + i_t \odot \tilde{c}_t \\
h_{t+1} &= o_t \odot \tanh(c_{t+1}), \\
\end{aligned}
\label{eq:lstm}
\end{equation}
where $W^{f,h}$, $W^{f,x}$, $W^{i,h}$, $W^{i,x}$, $W^{o,h}$, $W^{o,x}$, $b_f$, $b_i$, and $b_o$ represent the weight matrices and bias terms of the forget gate, input gate, and output gate, respectively. In addition, $W^{c,h}$, $W^{c,x}$ and $b_c$ denote the weight matrices and bias term of the candidate cell state, respectively; $\sigma$ and $\odot$ represent the logistic sigmoid function and element-wise multiplication operator, respectively.

\subsubsection{Confidence Assessment Module}
To evaluate the confidence of predictions in the roll motion forecasting task, a confidence assessment module composed of a parallel regression head and classification head is designed in the model architecture. Based on the temporal features extracted by the Bi-LSTM, this module not only predicts roll values but also outputs a discretized quantification of prediction reliability, thereby achieving a distributed representation of the prediction results.

The regression head is implemented as a fully connected layer that maps the extracted features to the roll prediction:

\begin{equation}
 \mathbf{R} = \text{Linear}(\text{Regression}(\mathbf{x})).
\end{equation}

In parallel, the classification head links the continuous roll values to a discrete probability distribution via interval-based discretization. Specifically, the range of roll values is partitioned into $N$ consecutive bins, each corresponding to a class index. The classification head projects the extracted features to logits and applies a Softmax function to obtain the probability distribution:

\begin{equation}
 \mathbf{P} = \text{Softmax}(\text{classification}(\mathbf{x})).
\end{equation}
Here, $\mathbf{R} \in \mathbb{R}^{B \times T \times C}$ and $\mathbf{P} \in \mathbb{R}^{B \times T \times N}$, where $B$ denotes the batch size, $T$ the number of time steps, $C$ the number of channels, and $N$ the number of discrete bins. For each time step $t$, the probability vector $\mathbf{P}_t = [P_1, P_2, \ldots, P_N]$ represents the model's confidence that the roll angle falls into each bin. Let $R_{\text{index}}$ denote the bin index associated with the regression output; the corresponding confidence score is defined as
\begin{equation}
\mathbf{P_{i}} = P_{R_{index}}.
\end{equation}

\subsection{Loss function and metric}

As depicted in Section~\ref{Prediction Paradigm}, roll forecasting is formulated as a multi-task learning problem with a regression branch for point prediction and a classification branch for confidence estimation. The classification branch outputs a probability distribution over $N$ discretized roll-angle bins; therefore, we supervise it using the Kullback--Leibler (KL) divergence:

\begin{equation}
\mathcal{L}_{C}(P^{(\mathrm{true})}\|P^{(\mathrm{pred})}) = \sum_{i=1}^{N} P_{i}^{(true)} \log \frac{P_{i}^{(true)}}{P_{i}^{(pred)}},
\end{equation}
where $P_{i}^{(\mathrm{true})}$ is the target distribution (e.g., generated by dynamic label smoothing), $P_{i}^{(\mathrm{pred})}$ is the predicted distribution, and $N$ is the number of discretized bins.

For the regression branch, the mean squared error (MSE) is employed as the loss function:
\begin{equation}
\mathcal{L}_{R}={\frac{\sum_1^n\left(y_i-\widehat{y}_i\right)^2}{n}},
\end{equation}
where $y_i$ denotes the ground-truth value and $\hat{y}_i$ denotes the predicted value.

To ensure that the point prediction is coherent with the confidence distribution, we further introduce a regression--classification consistency loss. Let $R_{\text{index}}$ denote the bin index associated with the regression output, and let $C_{\text{index}}$ denote the most confident bin from the classification branch. Since Argmax is non-differentiable, we adopt Soft-Argmax to obtain a differentiable alternative denoted as $\tilde{C}$ (expected bin index):

\begin{equation}
\tilde{C}_{\text{index}}=\sum_{i=1}^{N} i\cdot P_i.
\end{equation}

The consistency loss is then defined as
\begin{equation}
\mathcal{L}_{\mathrm{Cons}}=\frac{1}{n}\sum_{i=1}^{n}\left(\tilde{C}_{\text{index}}-R_{\text{index}}\right)^2.
\end{equation}

\begin{figure}[pos=htbp]
\centering
\includegraphics[width=0.6\linewidth]{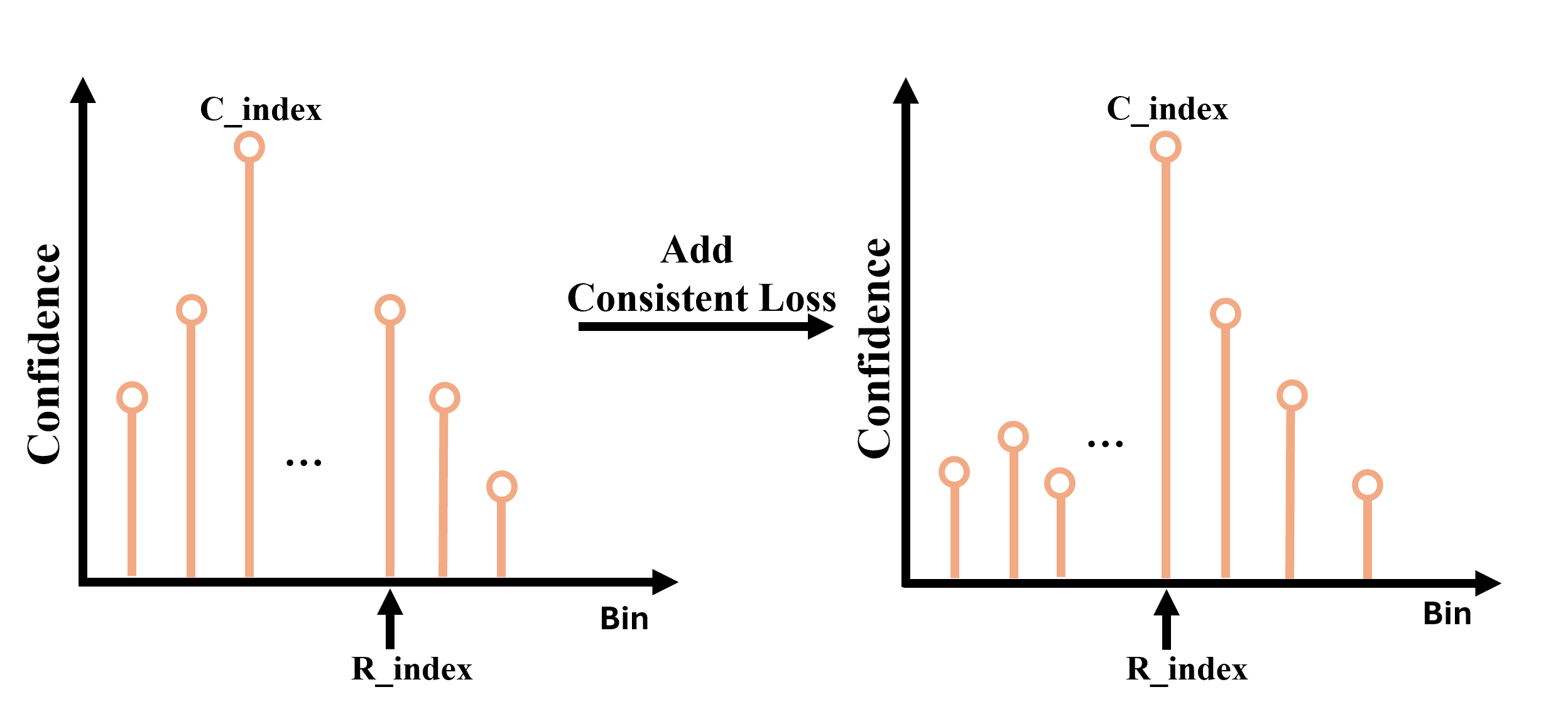}
\caption{Consistency effect}
\label{Consisit loss}
\end{figure}

The effect of introducing the consistency loss is illustrated in Fig.~\ref{Consisit loss}. The horizontal axis denotes the discretized bins, while the vertical axis represents the confidence that the prediction falls into each interval. As shown in the figure, in the left panel, the location of the maximum confidence from the classification head is inconsistent with the regression index, whereas in the right panel, after introducing the consistency loss, the alignment between the two is significantly improved.

The overall loss function $\mathcal{L}_{\mathrm{total}}$ is defined as follows, where \(\lambda_1\), \(\lambda_2\) and \(\lambda_3\) denote the weight coefficients for each loss component.
\begin{equation}
\mathcal{L}_{\mathrm{total}} = 
\lambda_1 \mathcal{L}_{\mathrm{C}} + 
\lambda_2 \mathcal{L}_{\mathrm{R}} + 
\lambda_3 \mathcal{L}_{\mathrm{Cons}}.
\end{equation}

To evaluate the model performance, both point prediction accuracy and interval prediction quality are considered. Point prediction accuracy is assessed using the root mean square error (RMSE) and mean absolute error (MAE), which measure the deviation between predicted values and actual observations. Lower RMSE and MAE values indicate higher prediction accuracy. Interval prediction quality is evaluated using the prediction interval coverage probability (PICP) and accumulated width deviation (AWD). PICP measures the proportion of true values that fall within the prediction intervals, with values closer to the target confidence level indicating higher interval reliability. AWD quantifies the relative deviation of observations lying outside the prediction intervals, where smaller AWD values indicate better interval prediction quality. The corresponding formulas are given as follows:

\begin{equation}
\mathrm{RMSE}=\sqrt{\frac{1}{n}\sum_{i=1}^{n}\left(y_i-\hat{y}_i\right)^2},
\end{equation}
\begin{equation}
\mathrm{MAE}=\frac{1}{n}\sum_{i=1}^{n}\left|y_i-\hat{y}_i\right|,
\end{equation}
\begin{equation}
\mathrm{PICP}=\frac{1}{n}\sum_{i=1}^{n}\mathbb{1}(l_i\le y_i\le u_i),
\end{equation}
\begin{equation}
\mathrm{AWD}=\frac{1}{n}\sum_{i=1}^{n}\left[\max(0,y_i-u_i)+\max(0,l_i-y_i)\right].
\end{equation}
 where $[l_i,u_i]$ denotes the prediction intervals,$l_i$ and $u_i$ are the lower bound and upper bound, $n$ is the total number of test samples, $y_i$ and $\hat{y}_i$ are the real value and predicted value.

\section{Dataset Analysis and Preprocessing}
\subsection{Dataset Analysis}
\label{Dataset Analysis}

To validate the proposed paradigm under realistic operating conditions, Xsens MTi-680G inertial navigation system were installed on a 7-meter-long high-speed vessel to record onboard navigation data at a sampling rate of 20 Hz. Given the average roll frequency of 0.5 Hz, the 20 Hz sampling yields considerable data redundancy. Equal-interval downsampling is adopted to reduce the sampling rate to 10 Hz for more efficient exploitation of historical roll information in model training.The main dimensional parameters of the platform are summarized in Table~\ref{tab:main_scale}.

\begin{table}[pos=htbp]
  \centering 
  \caption{USV dimension parameters}
  \label{tab:main_scale}     
  \begin{tabular}{l c r} 
    \toprule
    Dimensions & Symbol & Value \\
    \midrule
    Length between perpendiculars(m) & $L_{pp}$ & 7.5  \\
    Waterline breadth(m) & $B_{WL}$ & 2.6  \\
    Depth(m) & D & 1.61  \\
    Draft(m) & T & 0.54 \\
    Displacement($m^3$) & $\nabla$ & 3.62 \\
    Longitudinal center of gravity (m) & LGG & 2.45 \\
    Inertia moment of x axis/Beam at waterline & $i_{xx}/B$ & 0.35 \\ 
    \shortstack{Inertia moment of z axis/Length between the\\perpendiculars} & $i_{zz}/L_{pp}$ & 0.25  \\
    \bottomrule
  \end{tabular}
\end{table}

\begin{figure}[pos=htbp]
    \centering
    \begin{subfigure}{\linewidth}
        \centering
        \includegraphics[width=0.9\linewidth]{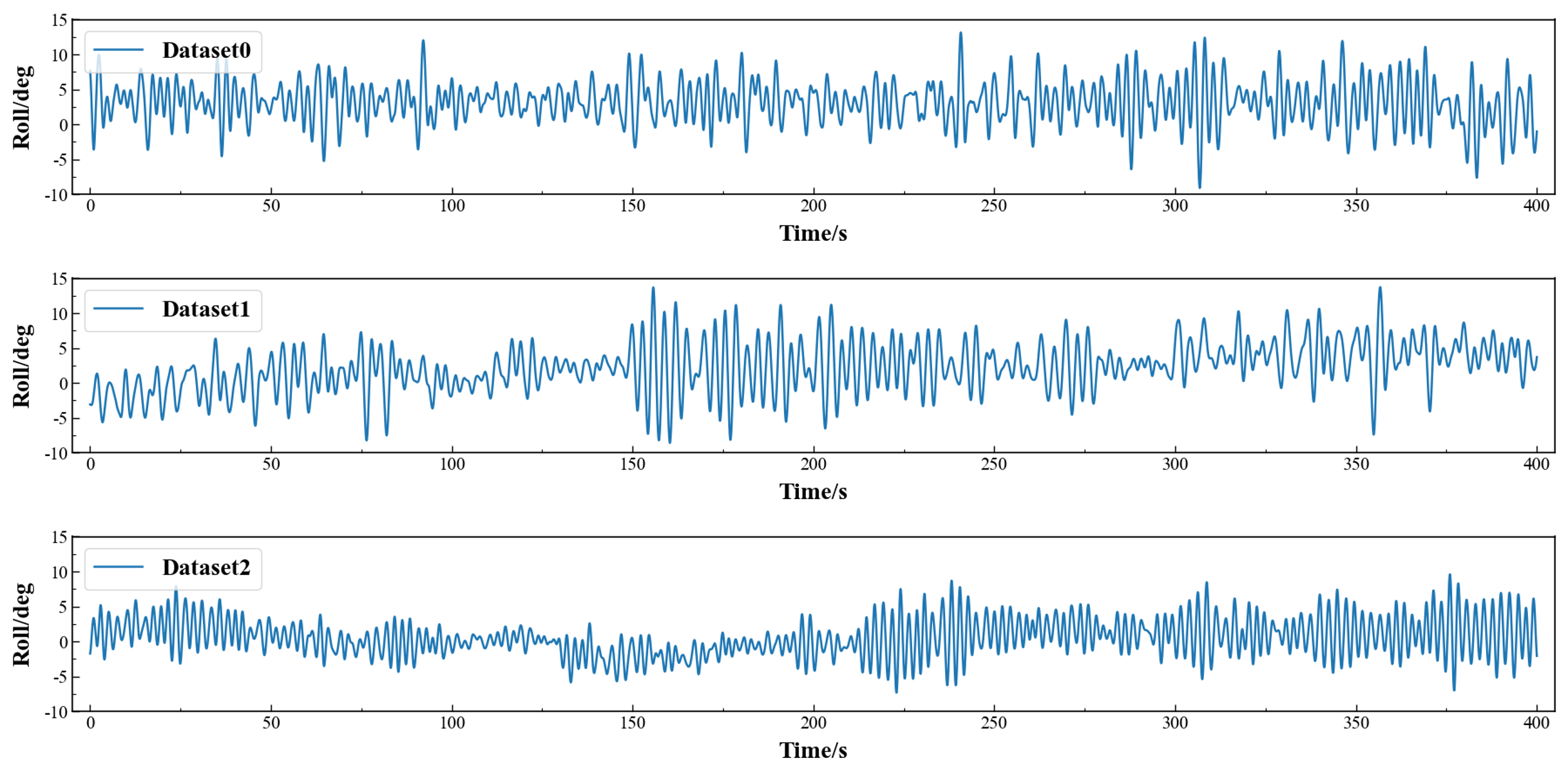}
        \caption{USV roll motion overview}
        \label{fig:roll_overview}
    \end{subfigure}
    \begin{subfigure}{\linewidth}
        \centering
        \includegraphics[width=0.5\linewidth]{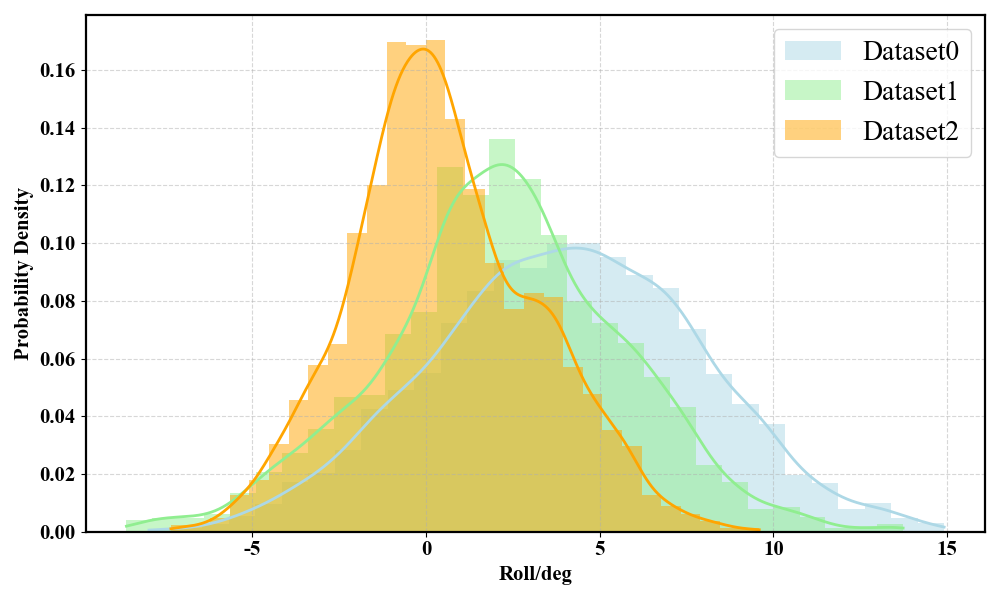}
        \caption{Distribution of roll}
        \label{fig:roll_hist_sub}
    \end{subfigure}
    \caption{Analysis of USV roll motion: (a) Time-series overview and (b) statistical distribution.}
    \label{fig:roll_combined}
\end{figure}

Fig.~\ref{fig:roll_combined} visualizes three roll datasets and their statistical distribution under different operation condition. And Table \ref{Multi-condition data} shows statistical description. As shown, roll angles mainly lie within $[-15^{\circ},\,15^{\circ}]$, which covers the dominant attitude variations under typical sea states. Extreme roll events beyond this range occur with very low probability and are usually associated with exceptional sea states or system instability, thus lacking statistical representativeness.  Accordingly, based on the empirical distribution of the measured data, the effective classification range for discretized roll bins used by confidence assessment head is set to $[-15^{\circ},\,15^{\circ}]$. 

The division of the 40 discrete roll bins is determined based on two primary considerations. 
From the perspective of practical USV navigation safety, a physical prediction resolution of $0.75^\circ$(corresponding to the 40 bins) is sufficiently granular to capture subtle attitude variations in roll motion, thereby providing an essential response margin for decision-making systems. And from the standpoint of model architecture design, excessively small bins would induce severe label sparseness that hinders model convergence, whereas overly large bins would fail to satisfy the required forecasting precision.

\begin{figure}[pos=b]
    \centering
    \begin{subfigure}{0.5\linewidth}
        \centering
        \includegraphics[width=\linewidth]{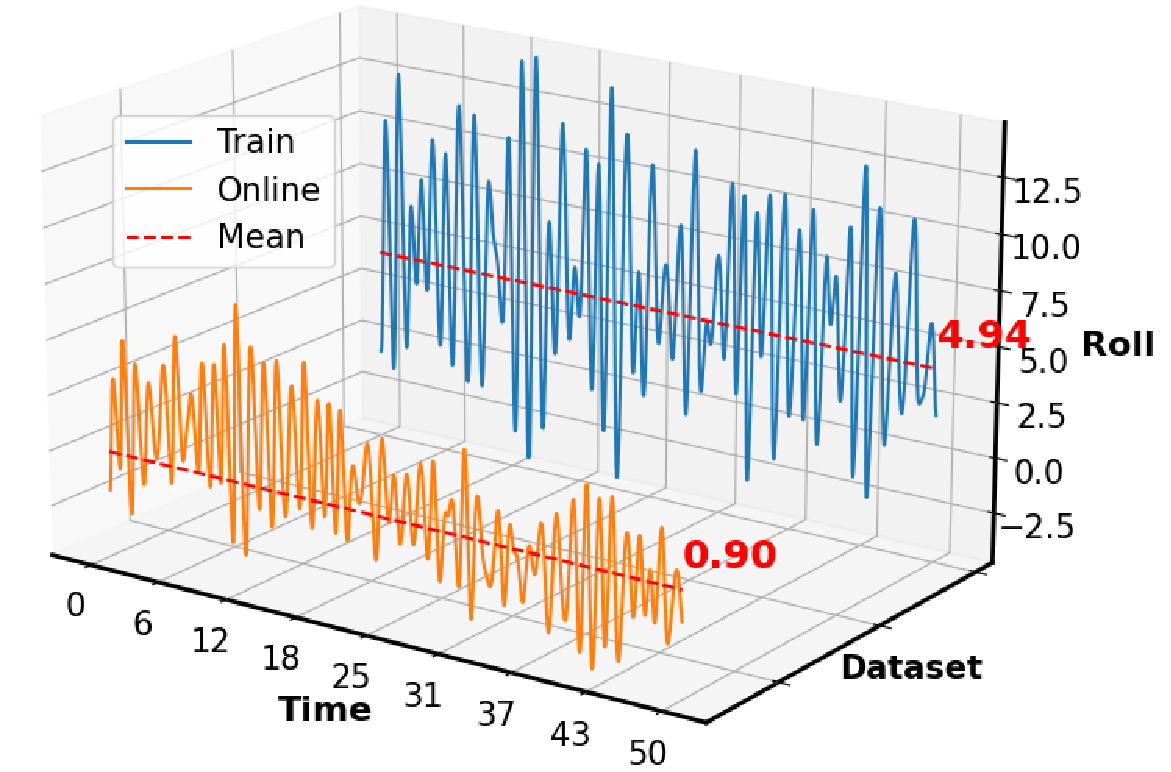}
        \caption{Train and online dataset distribution drift}
        \label{fig:dataset_drift}
    \end{subfigure}
    \begin{subfigure}{0.6\linewidth}
        \centering
        \includegraphics[width=0.9\linewidth]{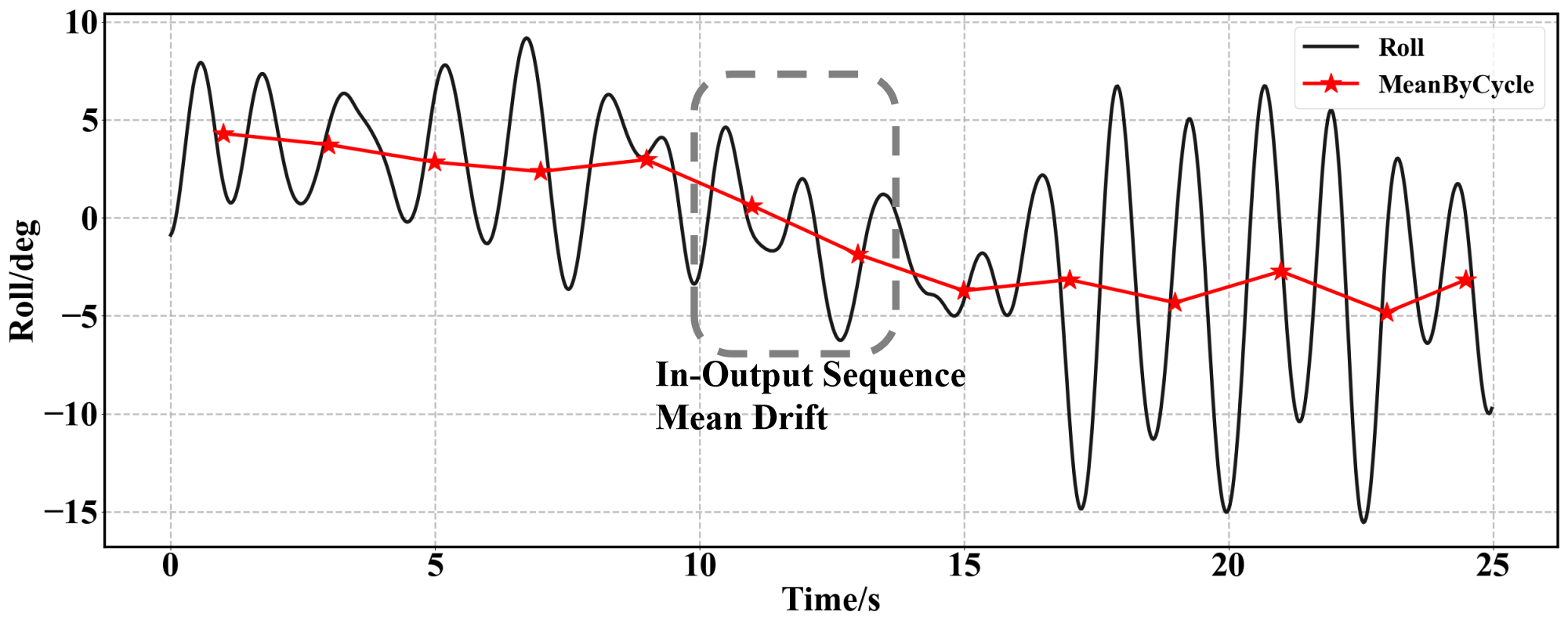}
        \includegraphics[width=0.9\linewidth]{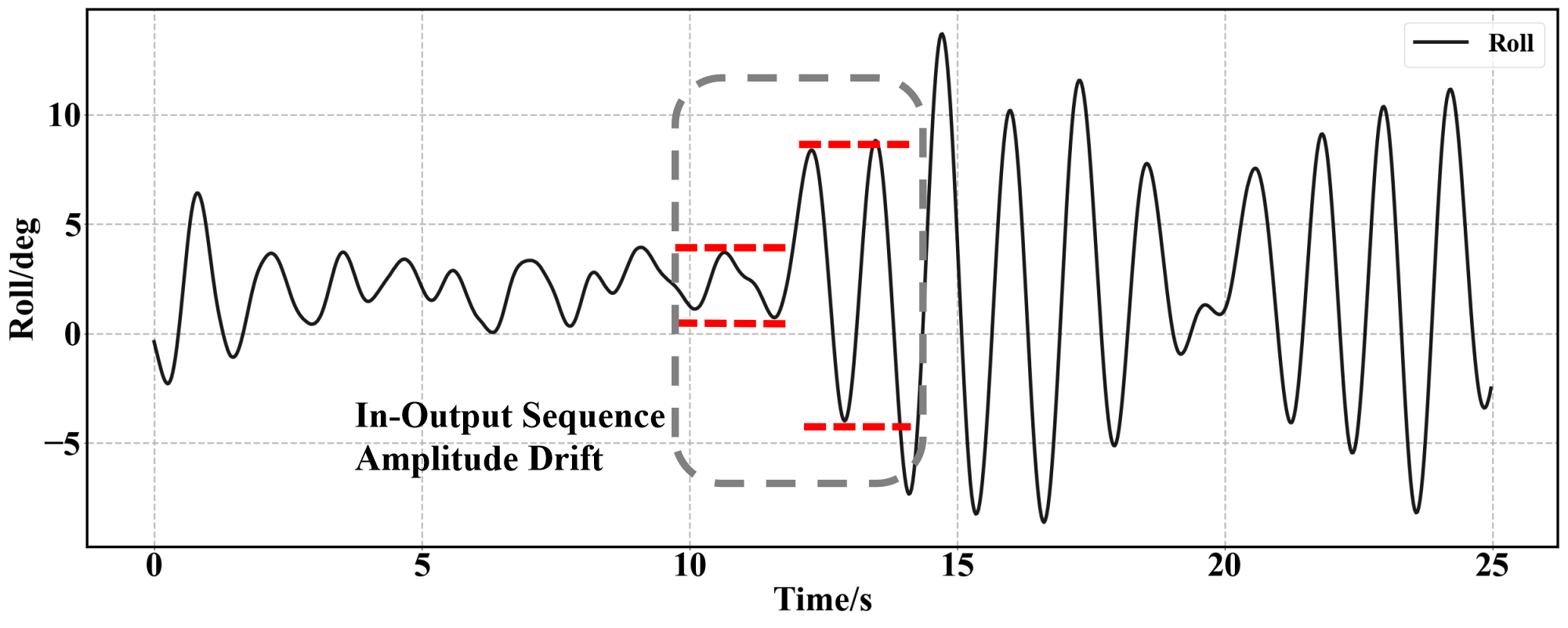}

        \caption{Input and output sequence distribution drift}
        \label{fig:window_drift}
    \end{subfigure}

    \caption{Distribution drift}
    \label{fig:distribution_drift}
\end{figure}

During navigation, the operating conditions of the USVs (such as speed, payload, and heading) and environmental conditions (such as wind and wave intensity, swell direction, and frequency) are continuously changing, their effects are ultimately manifested as distribution shifts.
In particular, two types of distribution shift are most detrimental to generalization: (i) the shift between training data and online data, and (ii) the shift between input and future sequences within a prediction window. These two scenarios are shown in Fig.~\ref{fig:distribution_drift}.

Considering that, we proposed the adaptive centralization method as depicted on Section \ref{section: Adaptive Centralization} and experiments will be conducted on Section \ref{section: Generalization Test of Adaptive Centralization} to compare it with No-standardization and Z-Score standardization. Specifically, Dataset \#0 is divided into training and validation sets with an 8:2 ratio, while Datasets \#1 and \#2 are reserved as independent test sets for performance evaluation.

\begin{table}[pos=htbp]
  \centering 
  \caption{Multi-condition dataset description}
  \label{Multi-condition data}     
  \resizebox{0.8\textwidth}{!}{ 
  \begin{tabular}{cccccccc} 
    \toprule
     Dataset & Sailing Condition & Sea State & Wave encounter & Sample Number & Mean(deg) & Max(deg) &Min(deg) \\    
    \midrule
    \#0 & Initial Condition & 4 &Beam Sea &28701 &3.151 & 14.660 & -13.550 \\
    \#1 & Altered Condition & 4 &Oblique Sea&6651  &2.201 & 13.730 &-8.621  \\
    \#2 & Altered Condition & 3 &Oblique Sea&8401  &0.547 & 9.590  &-7.340   \\
    \bottomrule
  \end{tabular}
  }
\end{table}

\subsection{Data Preprocessing}
\subsubsection{Sliding Window Method}
To fit the supervised learning setting, we convert the raw time series into input--output pairs using a sliding window strategy. Specifically, at each time step $t$, a historical observation sequence of length $d$ is selected as the model input, denoted as
\begin{equation}
X_t = [x_{t-d+1}, x_{t-d+2}, \ldots, x_t],
\end{equation}
and the subsequent future sequence of length $p$ is selected as the prediction target, denoted as
\begin{equation}
Y_t = [y_{t+1}, y_{t+2}, \ldots, y_{t+p}].
\end{equation}

\begin{figure}[pos=htb]
    \centering
    \includegraphics[width=0.6\linewidth]{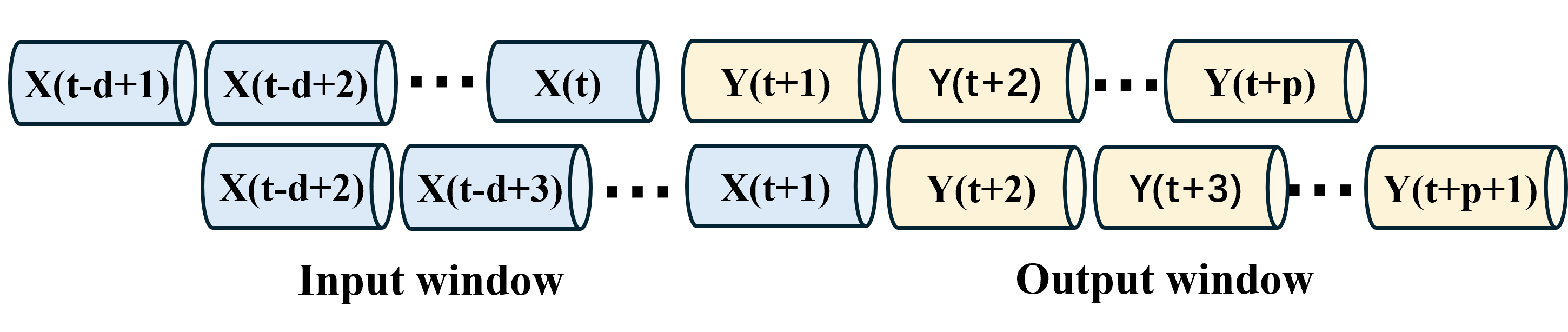}
    \caption{Sliding window method  }
    \label{fig:slide window}
\end{figure}

As illustrated in Fig.~\ref{fig:slide window}, the blue blocks represent the input sequence of the model, while the yellow blocks represent the output sequence to be predicted by the model.

The sliding window moves forward along the time series with a fixed stride of one sampling point. Specifically, after generating the sample pair $(X_t,Y_t)$, the window shifts by one time step to generate the next sample pair $(X_{t+1},Y_{t+1})$. Therefore, adjacent samples share $d-1$ historical observations, while their prediction targets are shifted forward by one step. This strategy enables efficient utilization of the continuous temporal information contained in the original roll motion data.

\subsubsection{Label Smoothing}

Since the proposed paradigm contains a confidence-assessment (classification) branch, the continuous roll angle must be converted into a supervision signal defined on discretized bins. A straightforward one-hot encoding assigns all probability mass to a single bin; however, this hard discretization introduces quantization error and produces unstable gradients for samples close to bin boundaries, which can further deteriorate confidence calibration.

Therefore, we employ a dynamic label smoothing strategy to generate a soft target distribution for training the classification branch (supervised by KL divergence). The label smoothing schematic as shown in Fig.~\ref{label_smooth}.
Given a continuous roll angle $x \in [Min, Max]$, the angle range is uniformly divided into $N$ bins. 
The bin width and the corresponding main bin index are calculated as:

\begin{figure}[pos=htb] \centering \includegraphics[width=0.6\linewidth]{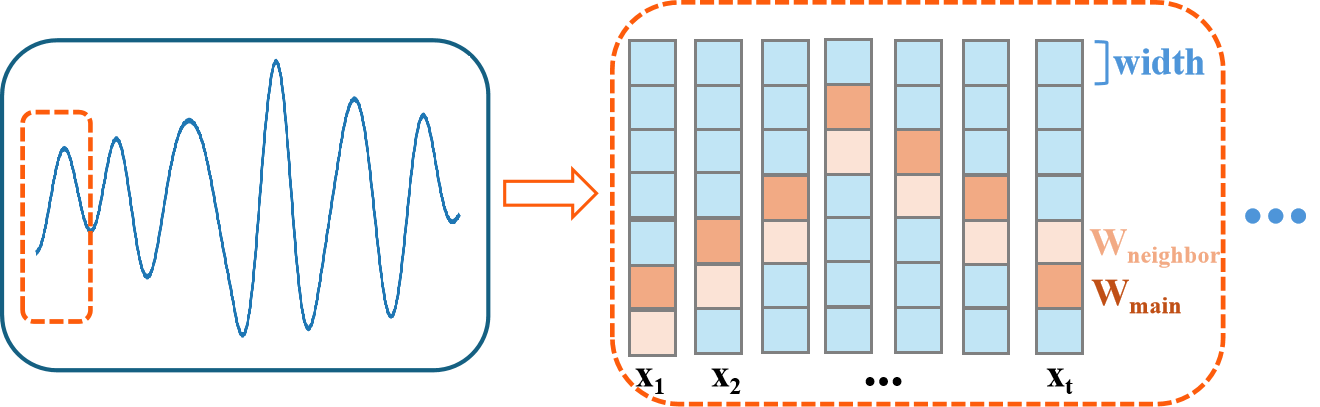} \caption{label smoothing} \label{label_smooth} 
\end{figure}

\begin{equation}
w=\frac{Max-Min}{N}
\end{equation}

\begin{equation}
i=\min\left(\left\lfloor\frac{x-Min}{w}\right\rfloor,N-1\right)
\end{equation}

where $i$ denotes the main bin containing the target value $x$. 
The center of the main bin is defined as:

\begin{equation}
c_i = \text{Min} + \frac{w}{2} + wi
\end{equation}

The normalized offset of $x$ relative to the bin center is given by:

\begin{equation}
\delta=\frac{x-c_i}{w}
\end{equation}

Instead of assigning a one-hot label, the proposed dynamic label smoothing strategy distributes the target probability between the main bin and its adjacent bin according to the relative position of $x$,

\begin{equation}
P_i=1-|\delta|
\end{equation}

\begin{equation}
P_{a}=|\delta|
\end{equation}
where $P_i$ and $P_a$ denote the main-bin probability and adjacent-bin probability. 
The probabilities of all remaining bins are set to zero:

\begin{equation}
P_j=0,\quad j\neq i,a .
\end{equation}

This design (i) preserves local continuity of roll angles, (ii) reduces sensitivity to discretization boundaries, and (iii) provides better-prepared targets for learning a well-calibrated confidence distribution.

\subsection{Hyperparameter Settings}

To ensure the reproducibility of the experimental results, the model parameters were configured in Table~\ref{model_param}. 
About the selection of kernel size, we performed a grid search over window size of 3, 5, and 7, and found that a window size of 3 achieved the lowest validation error. Therefore, k=3 was selected as the final configuration.
The remaining hyperparameters were set as follows: the random seed was fixed at [42, 123, 2026], the learning rate was set to 0.001, the batch size was set to 64, the number of training epochs was set to 100, and the Adam optimizer was adopted for model training.

\begin{table}[pos=htbp]
  \centering
  \caption{Model paremeters configuration}
  \label{model_param}
  \resizebox{0.6\textwidth}{!}{
  \begin{tabular}{l l l}
    \toprule
    Module & Parameter & Value \\
    \midrule
    \multirow{2}{*}{Adaptive module}
    & Kernel size of moving\_average  & 3 \\
    & Hidden dimension of linear & $[128,128]$ \\
    \midrule
    \multirow{4}{*}{Feature extract}
    & Bi-lstm layers & 3 \\
    & Hidden dimension of Bi-lstm & $[128,128]$ \\
    & Activation function & ReLU \\
    & Dropout rate & 0.3 \\
    \midrule
    \multirow{2}{*}{Regression head}
    & Hidden dimension of linear & $[128,20]$ \\
    & Activation function & ReLU \\
    \midrule
    \multirow{2}{*}{Classification head}
    & Hidden dimension of linear & $[128,40]$ \\
    & Activation function & ReLU \\
    & Temperature scaling & 0.5 \\
    \midrule
    Loss function & Loss weight ($\lambda_1,\lambda_2,\lambda_3$) & $1,0.5,1$ \\
    \bottomrule
  \end{tabular}
  }
\end{table}

\section{Result Analysis}
\subsection{Regression prediction performance}
\subsubsection{Generalization Test of Adaptive Centralization}
\label{section: Generalization Test of Adaptive Centralization}
Due to their small displacement and low rotational inertia, USVs usually exhibit a relatively high roll frequency. Therefore, we consider a short-horizon prediction with lengths of 10-20, which is sufficient to encompass approximately one roll cycle and capture the dominant roll dynamics.

\begin{figure}[pos=bp]
    \centering
    \begin{subfigure}[b]{0.48\linewidth}
        \centering
        \includegraphics[width=\linewidth]{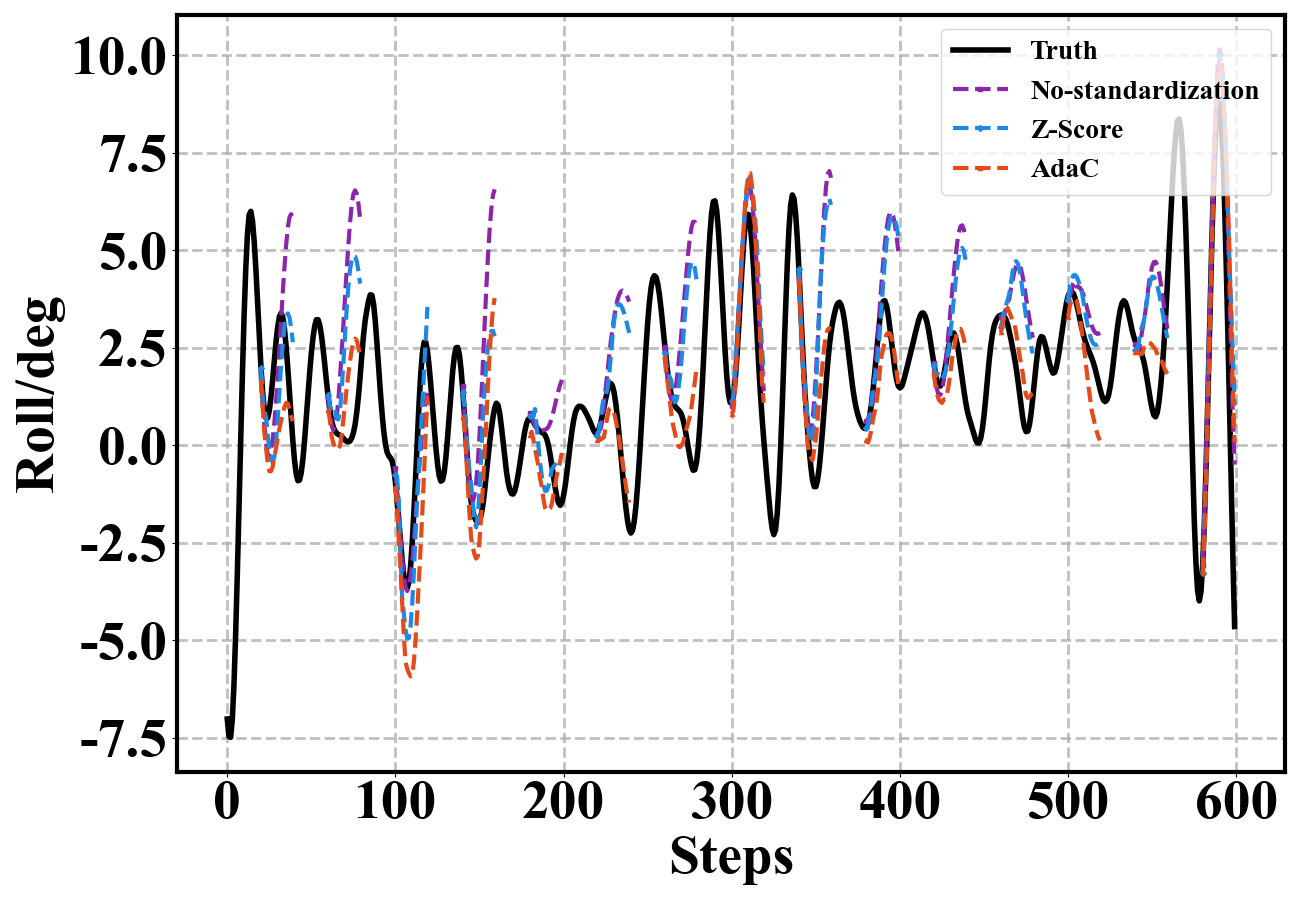}
        \caption{Dataset \#1}
        \label{fig:Generalization0a}
    \end{subfigure}
    \hfill
    \begin{subfigure}[b]{0.48\linewidth}
        \centering
        \includegraphics[width=\linewidth]{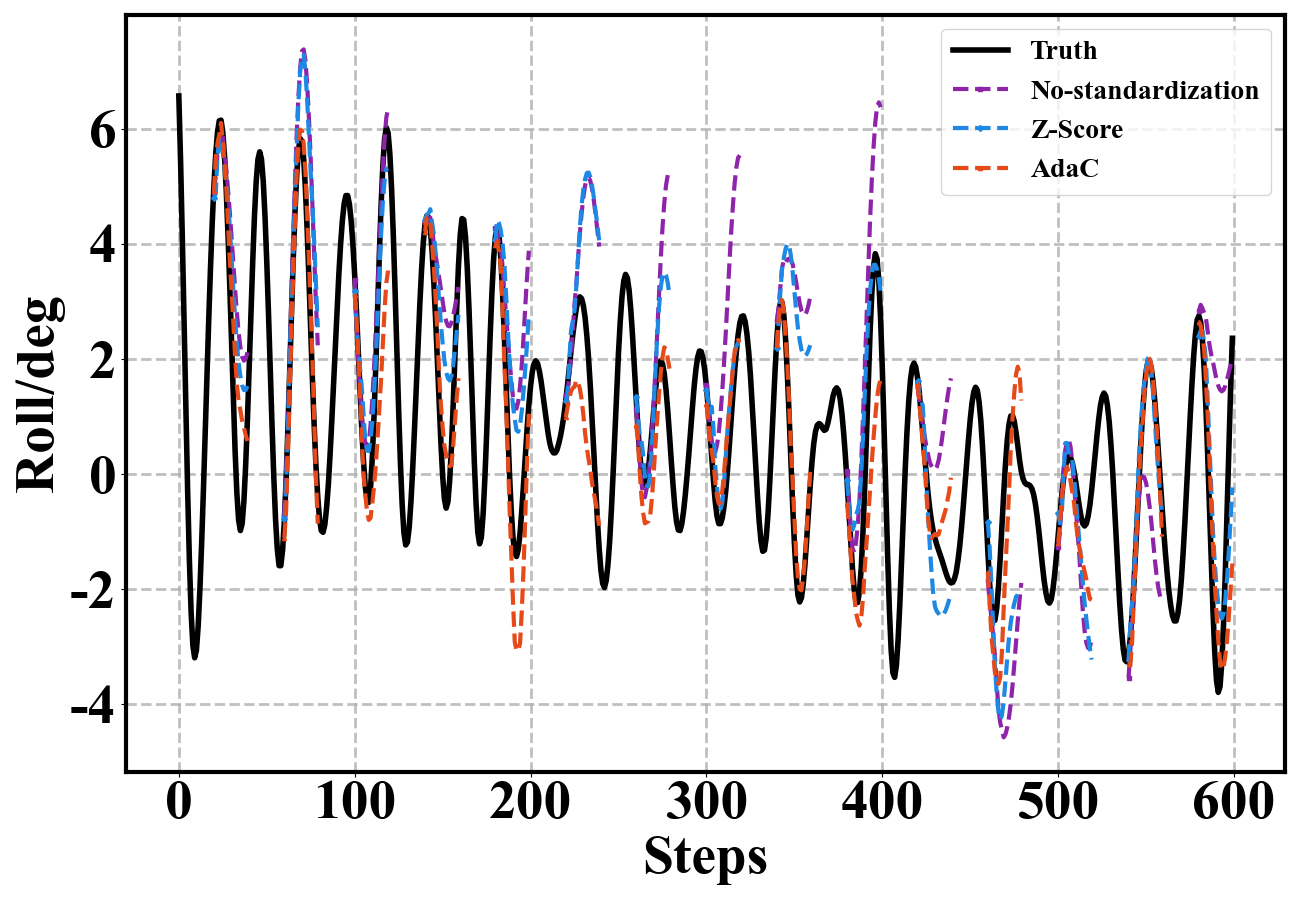}
        \caption{Dataset \#2}
        \label{fig:Generalization0b}
    \end{subfigure}
    
    \caption{Comparison between No-standardization, Z-Score and proposed AdaC}
    \label{fig:Generalization}
\end{figure}

\begin{figure}[pos=htbp]
\centering
\resizebox{0.7\linewidth}{!}{%
\begin{minipage}{\linewidth}

\begin{center}
    \normalsize\textbf{Dataset \#1}
\end{center}

\begin{subfigure}{\linewidth}
    \centering
    \includegraphics[width=\linewidth]{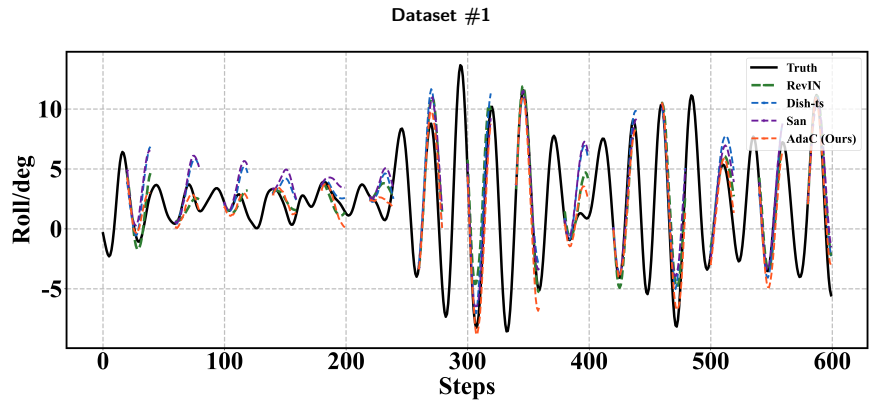}
    \caption{}
    \label{fig:generalization-a}
\end{subfigure}

\begin{subfigure}{\linewidth}
    \centering
    \includegraphics[width=\linewidth]{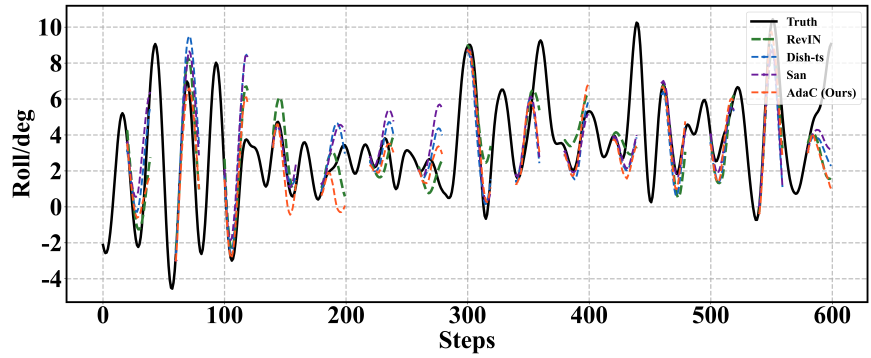}
    \caption{}
    \label{fig:generalization-b}
\end{subfigure}

\begin{center}
    \normalsize\textbf{Dataset \#2}
\end{center}

\begin{subfigure}{\linewidth}
    \centering
    \includegraphics[width=\linewidth]{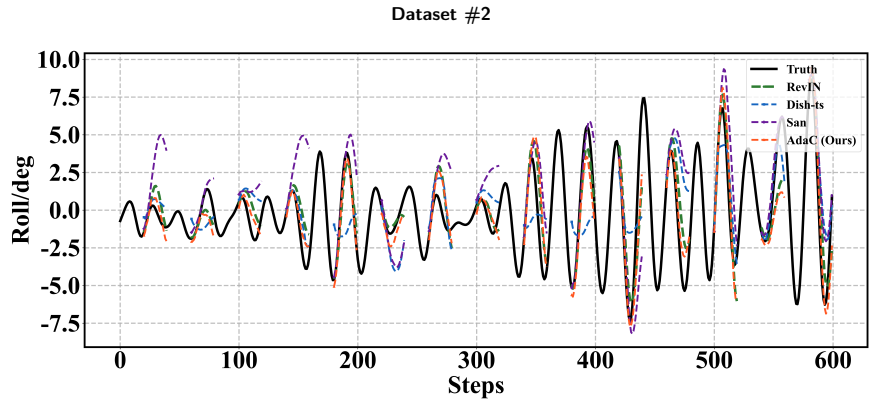}
    \caption{}
    \label{fig:generalization-c}
\end{subfigure}

\begin{subfigure}{\linewidth}
    \centering
    \includegraphics[width=\linewidth]{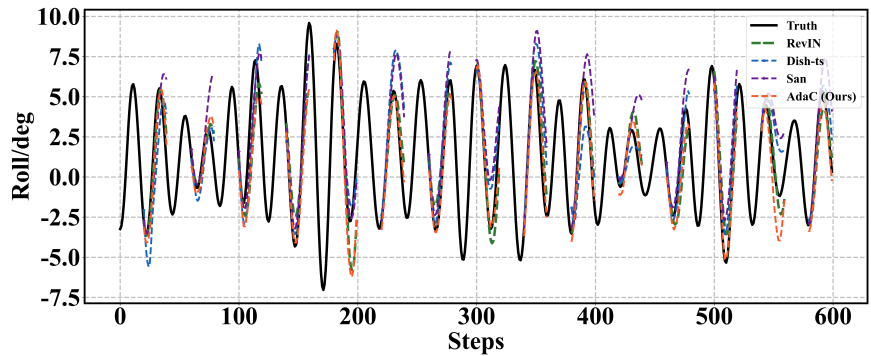}
    \caption{}
    \label{fig:generalization-d}
\end{subfigure}

\end{minipage}
}
\caption{Comparison among advanced normalization methods, including RevIN, Dish-ts, San, and the proposed AdaC.}
\label{fig:generalization}
\end{figure}

To evaluate the generalization benefit of the proposed adaptive centralization method, we conduct cross-condition experiments under different operating conditions based on the proposed reliability-aware prediction paradigm. Six preprocessing strategies are compared:  (1) No-standardization, (2) Z-Score standardization,(3) RevIN, (4) Dish-ts, (5)San, and (6) adaptive centralization(AdaC). Note that, in practical online prediction, the mean and variance required by Z-score statistics are not available, which are computed solely from the training set and applied to test conditions. 
Fig.~\ref{fig:Generalization} compares the basic preprocessing schemes, including no-standardization and Z-Score, while Fig.~\ref{fig:generalization} further compares advanced distribution adaptation methods, including RevIN, Dish-ts, San, and the proposed AdaC.
The quantitative results, evaluated using RMSE and MAE over multiple independent training runs, are summarized in Table~\ref{tab:RMSE_MAE}.

\begin{table}[b]
\centering
\caption{RMSE and MAE (\text{mean} $\pm$ std) of different preprocessing strategies under altered conditions}
\label{tab:RMSE_MAE}
\begin{tabular}{cccccccc}
\toprule
Dataset & Metric & No-standardization & Z-Score & RevIN & Dish-ts & San & AdaC (ours) \\
\midrule
\#1 & RMSE & 2.470 {\scriptsize$\pm$ 0.05} & 2.383 {\scriptsize$\pm$ 0.03} & 2.110 {\scriptsize$\pm$ 0.005} & 2.477
{\scriptsize$\pm$ 0.08} & 2.323 {\scriptsize$\pm$ 0.06} & \textbf{2.065 {\scriptsize$\pm$ 0.02}} \\
    & MAE  & 1.703 {\scriptsize$\pm$ 0.03} & 1.646 {\scriptsize$\pm$ 0.03} & 1.448 {\scriptsize$\pm$ 0.005} & 1.705
{\scriptsize$\pm$ 0.05} & 1.607 {\scriptsize$\pm$ 0.04} & \textbf{1.411 {\scriptsize$\pm$ 0.02}} \\
\midrule
\#2 & RMSE & 2.720 {\scriptsize$\pm$ 0.09} & 2.401 {\scriptsize$\pm$ 0.07} & 1.377 {\scriptsize$\pm$ 0.03} & 2.448
{\scriptsize$\pm$ 0.22} & 2.196 {\scriptsize$\pm$ 0.16} & \textbf{1.335 {\scriptsize$\pm$ 0.02}} \\
    & MAE  & 1.952 {\scriptsize$\pm$ 0.07} & 1.711 {\scriptsize$\pm$ 0.06} & 0.936 {\scriptsize$\pm$ 0.02} & 1.709
{\scriptsize$\pm$ 0.16} & 1.537 {\scriptsize$\pm$ 0.12} & \textbf{0.919 {\scriptsize$\pm$ 0.01}} \\
\bottomrule
\end{tabular}
\end{table}

To validate the capability of AdaC in mitigating distribution shifts, we compared it with the no-standardization strategy and the widely used Z-score  under varying operating conditions. As mentioned in Section ~\ref{Dataset Analysis}, the overall distributions of the online datasets (\#1 and \#2) are significant different from the training dataset. According to the RMSE and MAE results in Table~\ref{tab:RMSE_MAE}, AdaC reduces the prediction errors by approximately 13\% and 14\% on Dataset \#1, and 44\% and 46\% on Dataset \#2, respectively.
As shown in Fig.\ref{fig:Generalization0a} (200-500 steps) and Fig.\ref{fig:Generalization0b} (300-600 steps), significant input-output distribution shifts occur in these intervals. The prediction results of the non-standardization and Z-score normalization strategies exhibit large deviations and fail to effectively adapt to the changed distributions. In contrast, AdaC demonstrates strong robustness under distribution shifts. It not only alleviates the impact of distribution discrepancies but also maintains high prediction accuracy.

Considering the distribution shift between the training and online datasets, Table~\ref{tab:RMSE_MAE} compares  advanced distribution adaptation method.
On Dataset \#1, AdaC achieves the lowest RMSE and MAE among all methods. It outperforms RevIN, Dish-ts, and San, reducing RMSE by approximately 8\%-15\% and MAE by about 6\%-15\%.
On Dataset \#2, where baseline methods become more sensitive to distribution drift. AdaC achieves more pronounced improvements. Compared with all baseline methods, it reduces RMSE and MAE by 6\%--57\% and 3\%--59\%, respectively, while consistently achieving the lowest prediction errors.

Considering the amplitude shift between the input and output sequence, as shown in the Fig.~\ref{fig:generalization}. In Fig.~\ref{fig:generalization-a} (steps 200-320), Fig.~\ref{fig:generalization-c} (steps 100-200) and Fig.~\ref{fig:generalization-d}(steps 300-500), an input-output sequence distribution shift occurs. During this period, the Dish-ts and San predictions show a noticeable bias, and the RevIN method fails to accurately predict extreme values, only the adaptive centralization approach is able to maintain good fitting performance.
In Fig.~\ref{fig:generalization-b}(steps 200-320), significant mean shifts occur in both the input and output distributions. Adac maintains relatively stable performance compared with others.

Overall, adaptive centralization consistently achieves the best performance across both datasets, showing improved robustness under distribution shifts.

\subsubsection{Ablation study}
\begin{figure}[pos=htbp]
    \centering
    \begin{subfigure}[b]{0.35\linewidth}
        \centering
        \includegraphics[width=\linewidth]{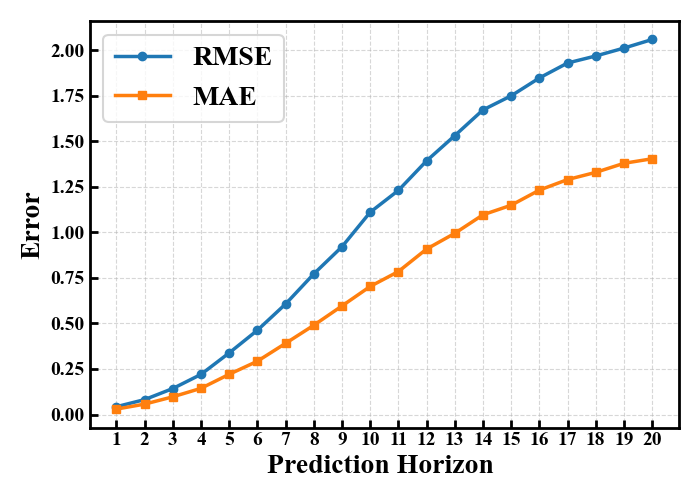}
        \caption{Dataset \#1}
        \label{fig:rmse}
    \end{subfigure}
    \hspace{0.03\linewidth}
    \begin{subfigure}[b]{0.35\linewidth}
        \centering
        \includegraphics[width=\linewidth]{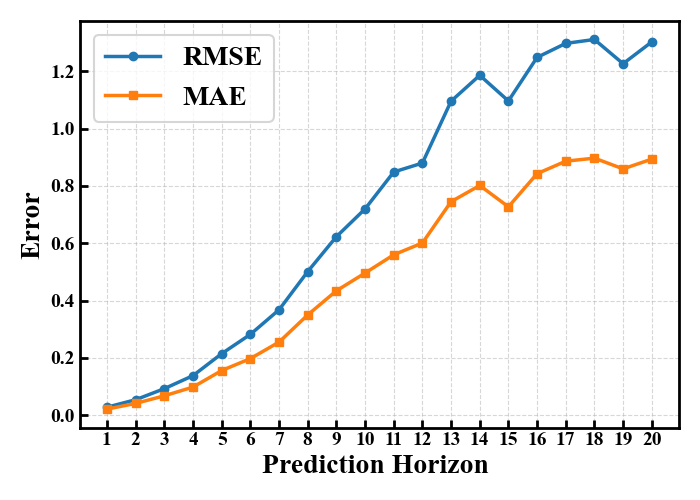}
        \caption{Dataset \#2}
        \label{fig:mae}
    \end{subfigure}

    \caption{Prediction error analysis under different forecasting horizons.}
    \label{fig:error_analysis}
\end{figure}

To evaluate the regression prediction performance of the model under different input and output lengths, an ablation study on the input-output window configurations was conducted. As shown in Fig. ~\ref{fig:error_analysis}, the RMSE and MAE gradually increase with the expansion of the prediction horizon.

\begin{figure}[pos=htbp]
    \centering
    \begin{subfigure}[b]{0.35\linewidth}
        \centering
        \includegraphics[width=\linewidth]{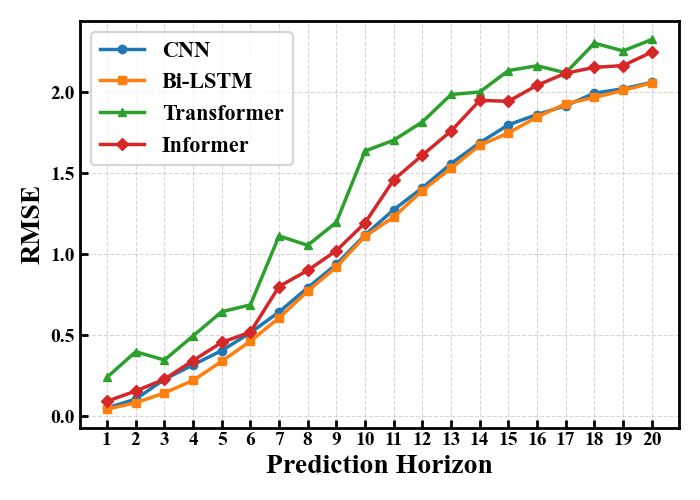}
        \caption{Dataset \#1}
        \label{fig:model_1}
    \end{subfigure}
    \hspace{0.03\linewidth}
    \begin{subfigure}[b]{0.35\linewidth}
        \centering
        \includegraphics[width=\linewidth]{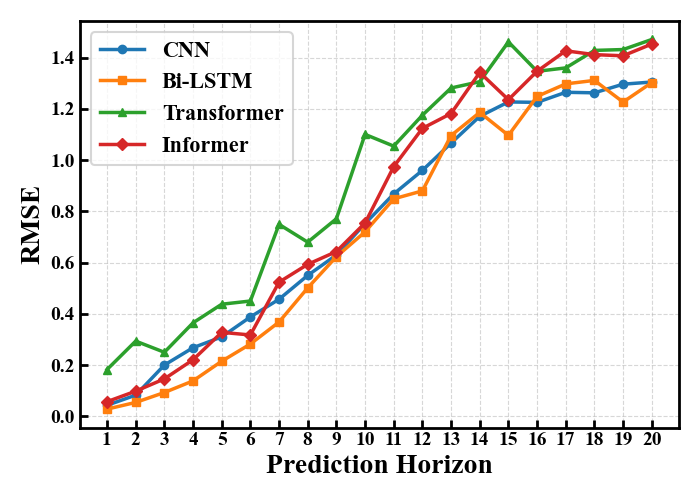}
        \caption{Dataset \#2}
        \label{fig:model_2}
    \end{subfigure}
    \caption{Ablation study on different backbone feature extractors}
    \label{fig:model_ablation}
\end{figure}

To verify the model-agnostic property of the proposed method, ablation experiments were conducted by replacing the backbone feature extractor. Specifically, the original Bi-LSTM backbone was replaced with CNN, Transformer, and Informer, while keeping other components unchanged. 
As shown in Fig.~\ref{fig:model_ablation}, the RMSE results under different backbone feature extraction networks are presented to evaluate the model-agnostic property of the proposed framework. It can be observed that the Bi-LSTM backbone achieves the lowest overall RMSE. Compared with Transformer-based models, it significantly reduces the prediction error, while showing a clear advantage over CNN-based models under shorter prediction horizons. 

\subsection{Reliability-aware Prediction Performance}
\subsubsection{Step-wise confidence Assessment}

\begin{figure}[pos=htbp]
    \centering
    \begin{subfigure}{0.49\linewidth}
        \centering
        \includegraphics[width=\linewidth]{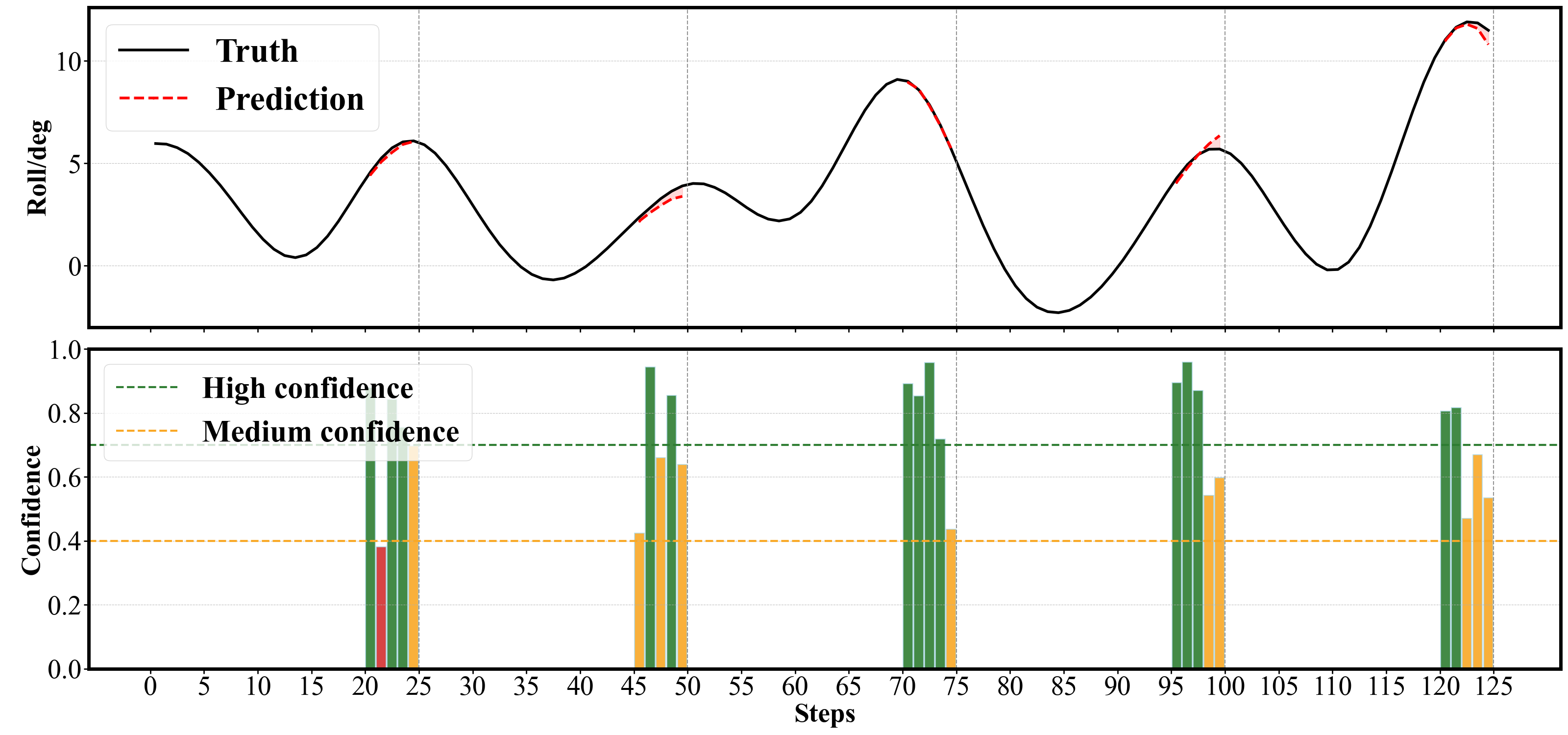}
        \caption{Confidence assessment for 5-steps prediction}
        \label{fig:5confidence}
    \end{subfigure}
    \hfill 
    \begin{subfigure}{0.49\linewidth}
        \centering
        \includegraphics[width=\linewidth]{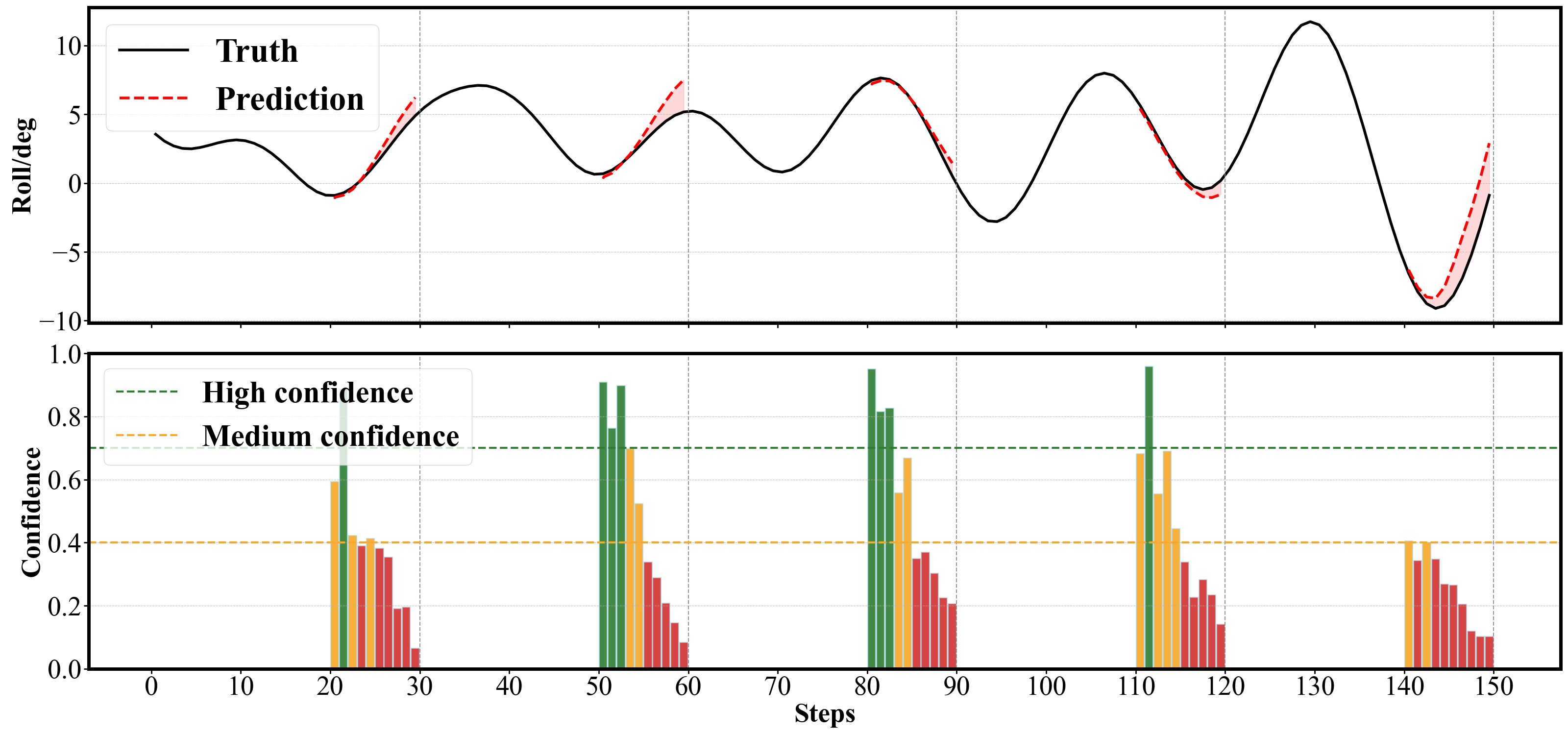}
        \caption{Confidence assessment for 10-steps prediction}
        \label{fig:10confidence}
    \end{subfigure}

    \vspace{2mm} 

    \begin{subfigure}{0.49\linewidth}
        \centering
        \includegraphics[width=\linewidth]{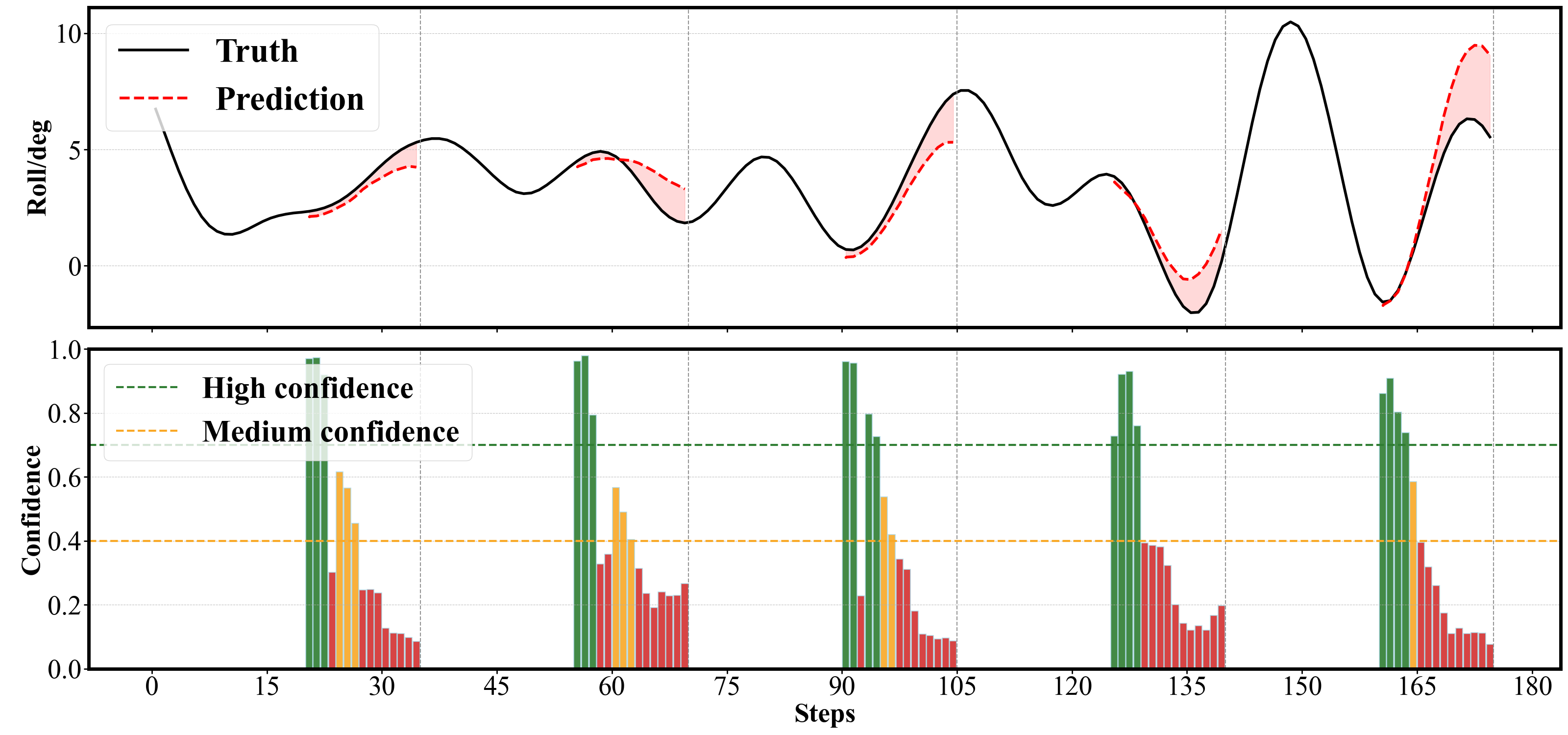}
        \caption{Confidence assessment for 15-steps prediction}
        \label{fig:15confidence}
    \end{subfigure}
    \hfill
    \begin{subfigure}{0.49\linewidth}
        \centering
        \includegraphics[width=\linewidth]{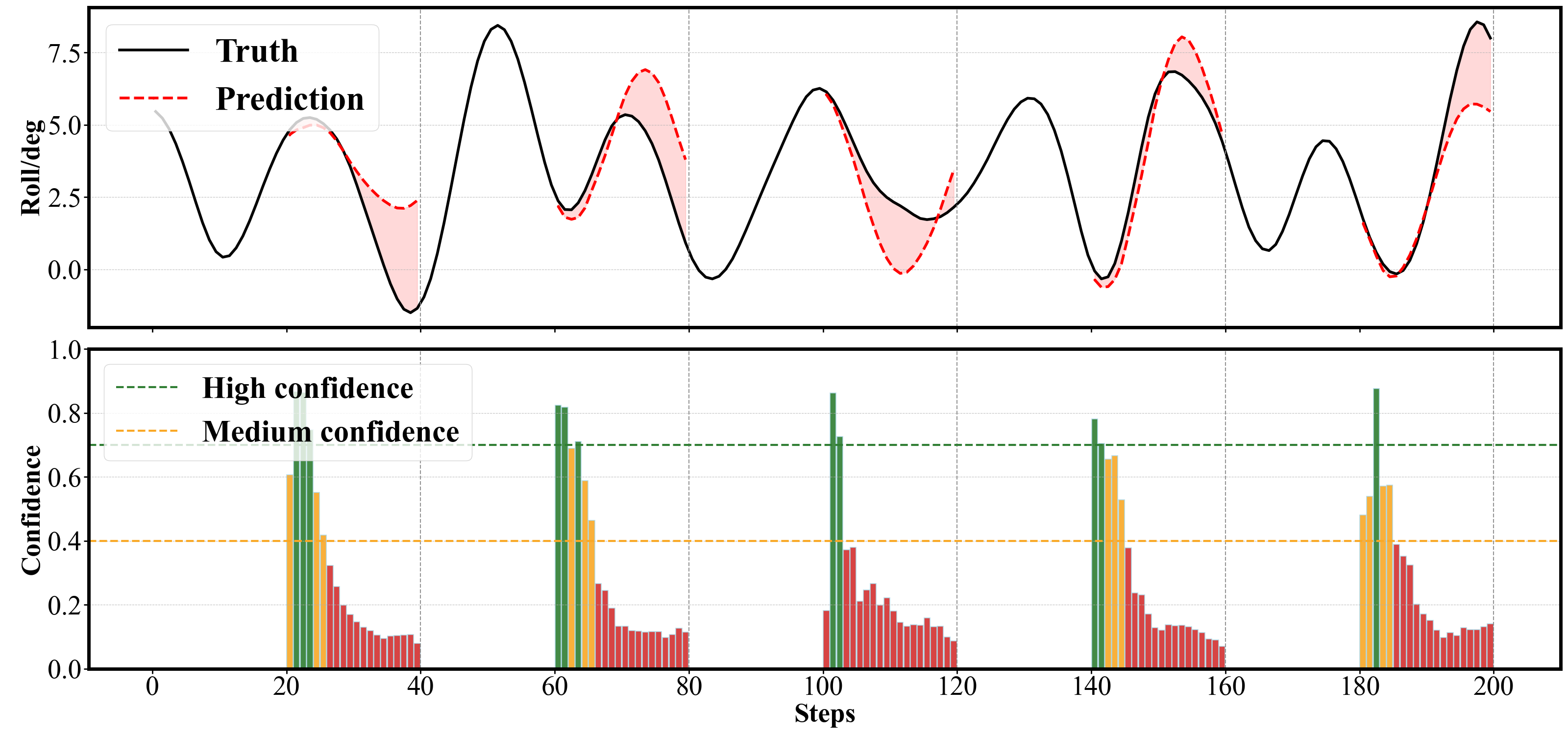}
        \caption{Confidence assessment for 20-steps prediction}
        \label{fig:20confidence}
    \end{subfigure}
    
    \caption{Confidence assessment over each step of prediction}
    \label{fig:confidence}
\end{figure}

By applying the reliability-aware prediction paradigm proposed in this paper, the reliability of the prediction results at each step can be quantitatively evaluated. The input-output settings are 20-5, 20-10,20-15 and 20-20, as shown in Fig.~\ref{fig:confidence}.
The upper plot shows the comparison between multi-step predictions and ground truth values, while the lower bar charts visualizes the step-wise confidence scores, where the bar height indicates the step-wise confidence level.

In the lower bar charts, confidence scores are divided into three levels using two thresholds: 0.7-1 represents the high-confidence interval (green), 0.4-0.7 represents the medium-confidence interval (yellow), and 0-0.4 represents the low-confidence interval (red). These thresholds are empirically chosen for visualization. 
From the overall distribution trend, confidence generally decreases as the prediction horizon increases, which is consistent with the general observation that uncertainty accumulates over time in multi-step prediction. 
This trend confirms the expected relationship between confidence and horizon length, supporting the rationality of the confidence estimates generated by the proposed regression--classification joint training framework.

More importantly, the model’s high-confidence performance in the early stages confirms its short-term prediction stability; once the confidence falls below the low-confidence threshold ($<0.4$), the regression results simultaneously exhibit significant deviations from the ground truth. This observation indicates that the confidence score can serve as an actionable indicator for identifying when predictions transition from \emph{reliable} to \emph{unreliable}, thereby providing a practical basis for risk-aware decision-making and safety-oriented prediction truncation in real-world USV applications.

\subsubsection{Interval Confidence Assessment}
Building upon the step-wise confidence assessment described in the previous section, we further construct prediction intervals using the discrete probability mass function produced by the classification head. Specifically, for each forecasting step, the classification head outputs a categorical distribution over $M$ discretized roll bins, denoted as $\{p_{t,1},p_{t,2},\ldots,p_{t,M}\}$, where $p_{t,i}$ represents the predicted probability mass associated with the $i$-th roll bin.

Given a nominal confidence level $\alpha \in (0,1)$ (e.g., 0.80, 0.90, 0.95), we interpret $\alpha$ as a quantile threshold that defines a high-probability region in the predictive distribution.Then we form a high-quality (HQ) confidence interval by selecting the most probable bins with the smallest possible total mass that reaches $\alpha$. Concretely, the bin probabilities $\{p_{t,i}\}$ are sorted in descending order, and their cumulative sum is computed. Let $\mathcal{S}_t(\alpha)$ denote the smallest set of bin indices whose cumulative probability first exceeds $\alpha$:
\begin{equation}
\sum_{i\in \mathcal{S}_t(\alpha)} p_{t,i} \ge \alpha.
\end{equation}
Finally, the lower and upper bounds of the prediction interval are obtained by taking the minimum and maximum roll values (or bin centers/edges) among the selected bins:
\begin{equation}
\mathrm{PI}_t(\alpha)=\big[\, \min_{i\in \mathcal{S}_t(\alpha)} b_i,\ \max_{i\in \mathcal{S}_t(\alpha)} b_i \,\big],
\end{equation}
where $b_i$ denotes the representative roll value of the $i$-th bin (e.g., bin center), and $\mathrm{PI}_t(\alpha)$ is the resulting prediction interval at time step $t$.

Huang \citep{huang2026deterministic} proposed two probabilistic baselines for roll forecasting, namely the Normal-Distribution (ND) and $t$-Distribution (TD) models. These models assume that the future roll response $y$ follows either a Gaussian or Student's $t$-distribution:
\begin{align}
y &\sim \mathcal{N}(\mu, \sigma^2), \label{eq:normal_dist} \\
y &\sim t_{\nu}(\mu, \sigma^2), \label{eq:t_dist}
\end{align}
where $\mu$, $\sigma^2$, and $\nu$ denote the mean, variance, and degrees of freedom, respectively.

To capture temporal dependencies, an LSTM-based encoder-decoder architecture is employed to extract hidden features $\mathbf{h}_t$ from the historical sequence $\mathbf{X}_t$:
\begin{equation}
\mathbf{h}_t = f_{\mathrm{LSTM\_encoder}}(\mathbf{X}_t).
\end{equation}

The decoder then outputs the estimated parameters $\hat{\mu}, \hat{\sigma}$ governing the target distribution $\mathcal{D}$. Multi-step autoregressive forecasting is subsequently performed via Monte Carlo sampling:
\begin{equation}
\hat{\mu}, \hat{\sigma} = f_{\mathrm{LSTM\_decoder}}(\mathbf{h}_t).
\end{equation}
\begin{equation}
\hat{y}^{(m)} \sim \mathcal{D}\left(\hat{\mu}, \hat{\sigma}\right), \qquad m = 1, \dots, M,
\end{equation}
where $M$ is the number of Monte Carlo samples. For a given significance level $\alpha$, the $(1-\alpha)$ prediction confidence interval $[L, U]$ is directly derived from the parametric distribution:
\begin{equation}
L = \hat{\mu} - z_{1-\alpha/2} \cdot \hat{\sigma}, \qquad 
U = \hat{\mu} + z_{1-\alpha/2} \cdot \hat{\sigma}.
\end{equation}

\begin{figure}[pos=htbp]
    \centering
    \begin{tikzpicture}[baseline=-0.5ex, scale=1]
    \draw[black, solid, line width=1pt] (0,0) -- (0.8,0);
    \node[right] at (1.0,0) {Target};
    \filldraw[fill=cyan!20!white, draw=cyan!70!black, line width=0.3pt] 
          (3,0.25) rectangle (3.6,-0.25);
    \node[right] at (3.8,0) {ND PI};
    \filldraw[fill=red!20!white, draw=red!50!black, line width=0.3pt] 
        (6,0.25) rectangle (6.6,-0.25);
    \node[right] at (6.8,0) {TD PI};
    \filldraw[fill=orange!15!white, draw=orange!60!black, line width=0.3pt] 
        (9,0.25) rectangle (9.6,-0.25);
    \node[right] at (9.8,0) {HQ PI};
    \end{tikzpicture}
    \par\vspace{0.5em}
    \begin{subfigure}{0.3\textwidth}  
        \centering
        \includegraphics[width=\linewidth]{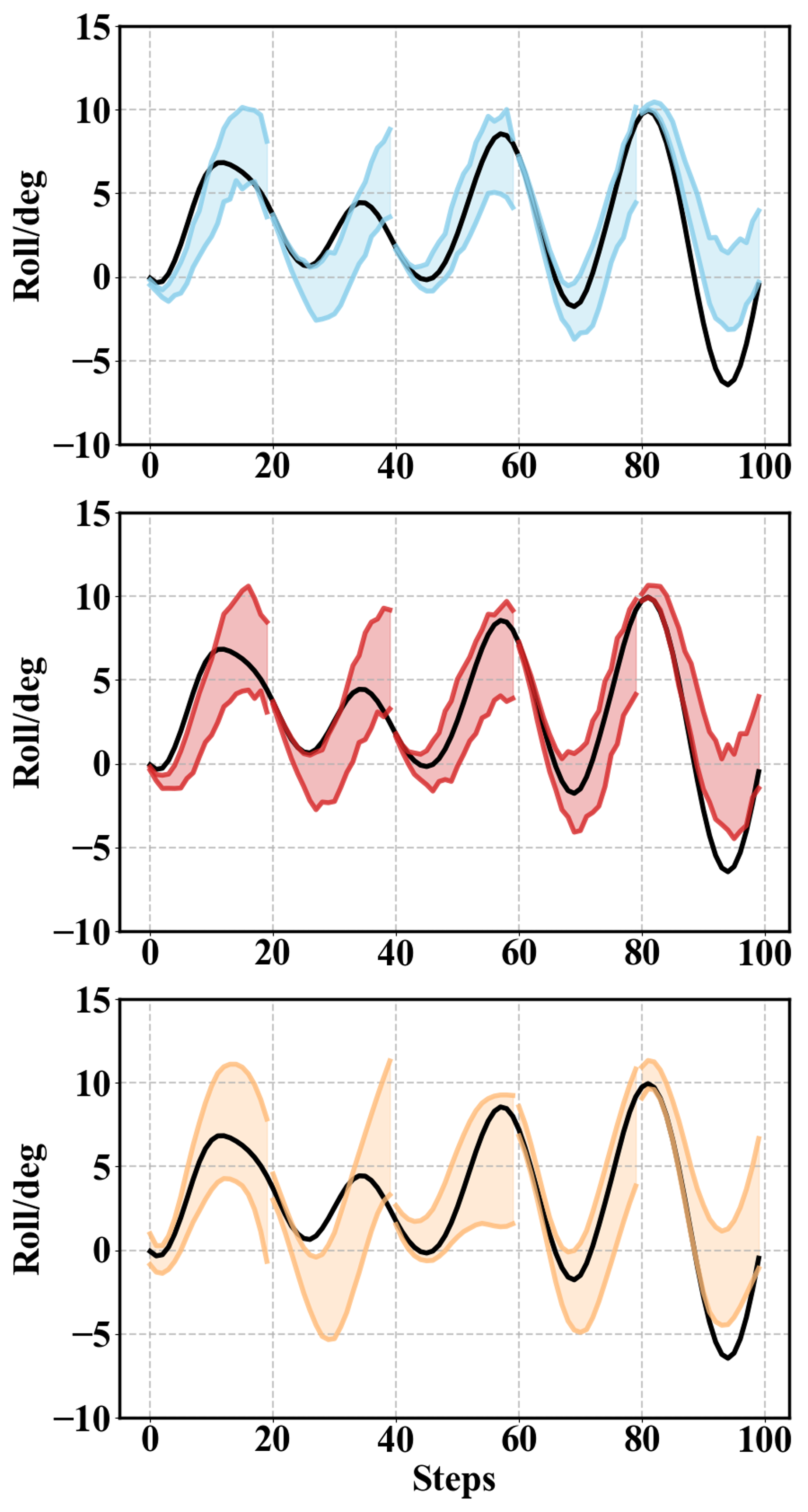} 
        \caption{80\% Confidence Level}
        \label{fig:a}
    \end{subfigure} 
    \begin{subfigure}{0.3\textwidth}
        \centering
        \includegraphics[width=\linewidth]{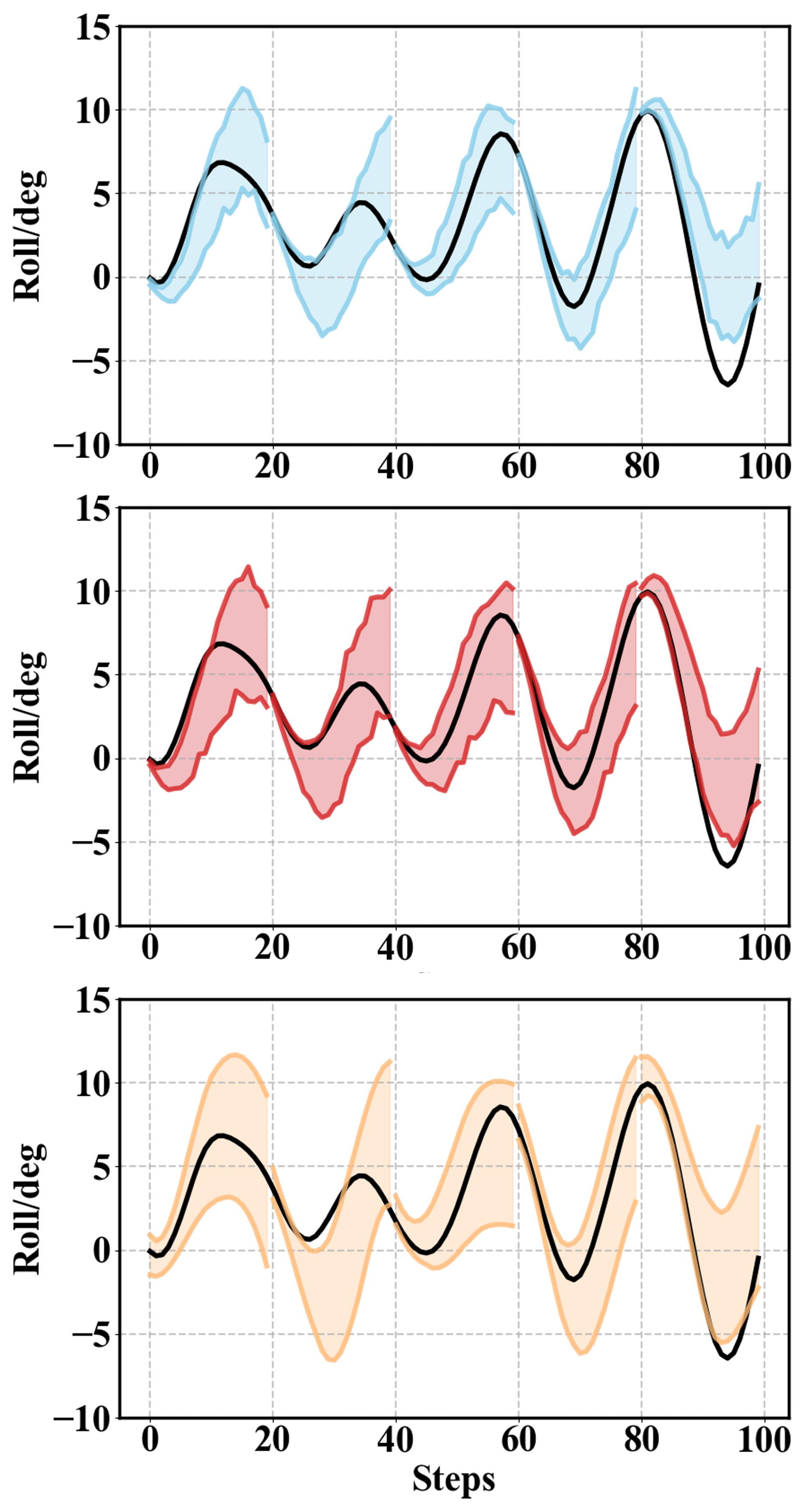}
        \caption{90\% Confidence Level}
        \label{fig:b} 
    \end{subfigure}
        \begin{subfigure}{0.3\textwidth}
        \centering
        \includegraphics[width=\linewidth]{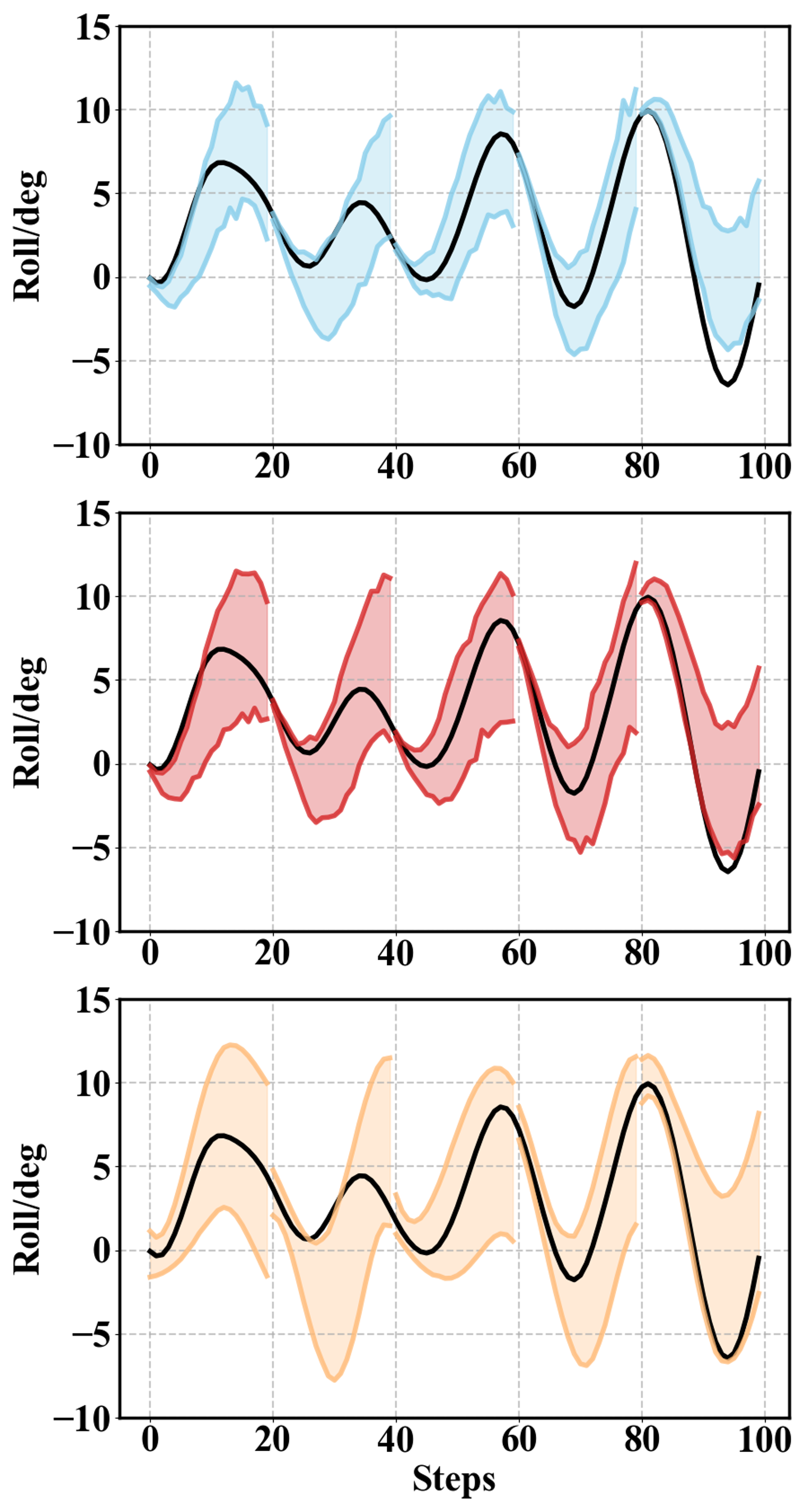}
        \caption{95\% Confidence Level}
        \label{fig:c} 
    \end{subfigure}
    \caption{The comparison of confidence interval prediction at different confidence levels}
    \label{fig:Comparison of PI}
\end{figure}
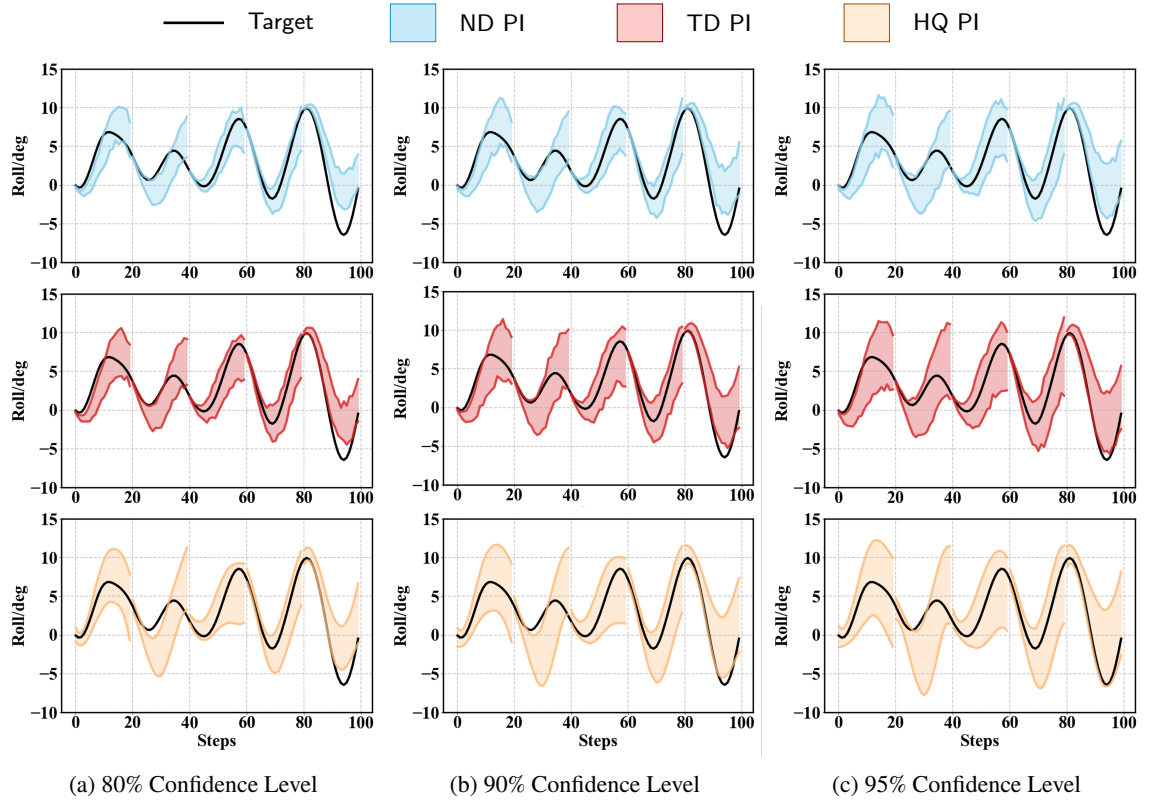

\begin{figure}[pos=htbp]
\centering
\includegraphics[width=\linewidth]{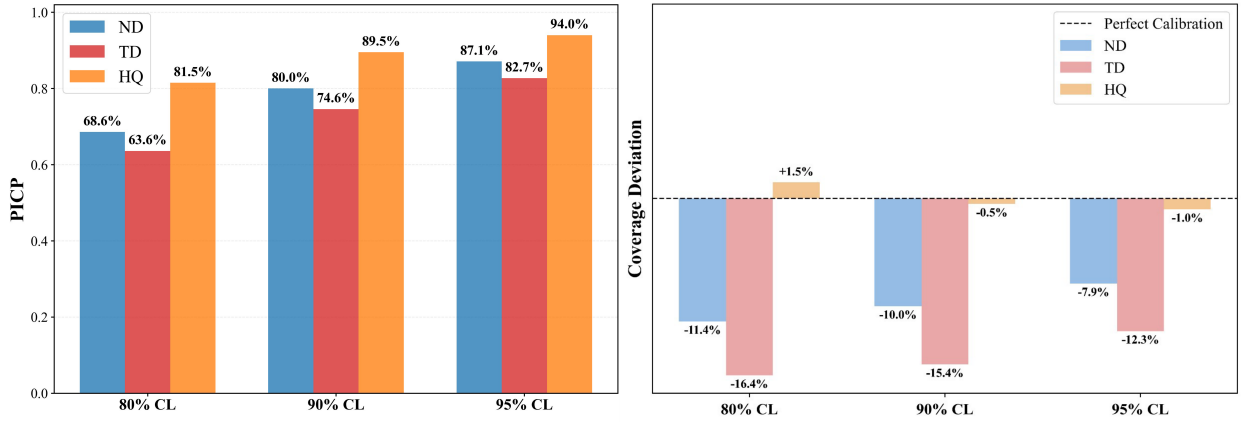}
\caption{Comparison of prediction interval coverage probability (PICP) for ND, TD, and the proposed HQ method under different nominal confidence levels (80\%, 90\%, and 95\%).}
\label{fig:PICP}
\end{figure}

\begin{figure}[pos=htbp]
\centering
\includegraphics[width=0.5\linewidth]{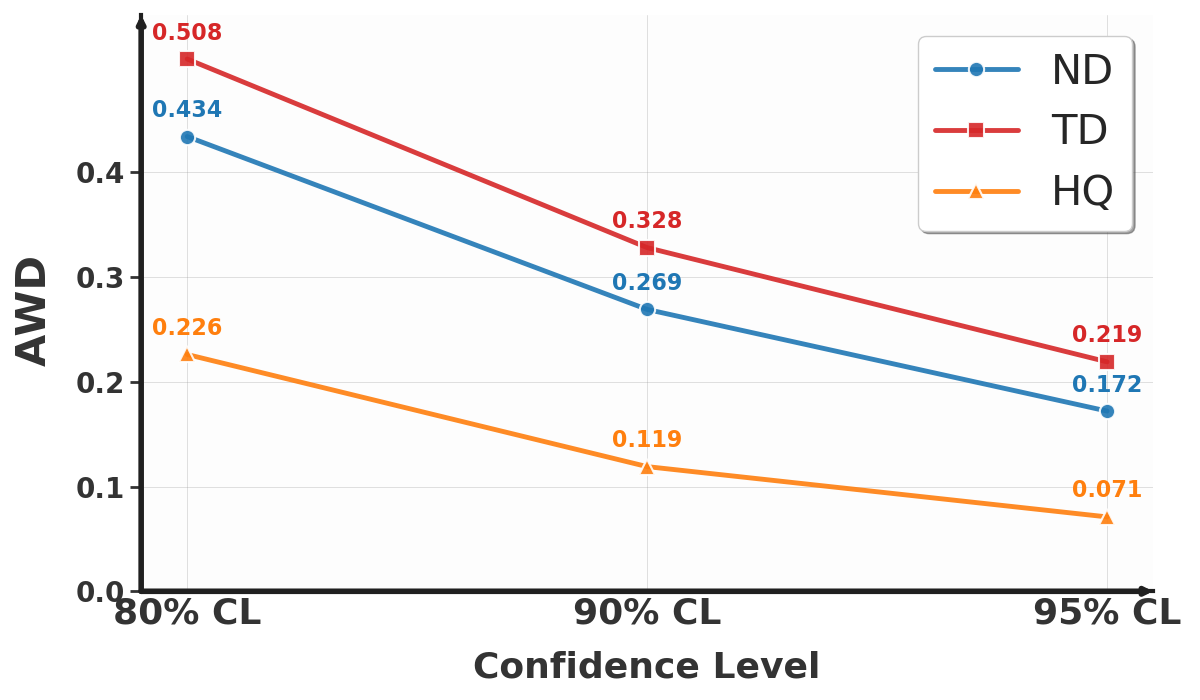}
\caption{Comparison of average width deviation (AWD) for ND, TD, and the proposed HQ method under different nominal confidence levels (80\%, 90\%, and 95\%).}
\label{fig:AWD}
\end{figure}

In this work, ND and TD are adopted as baseline methods to validate the effectiveness of the proposed HQ interval construction. Fig.~\ref{fig:Comparison of PI} compares the prediction intervals generated by ND, TD, and HQ at the 80\%, 90\%, and 95\% nominal confidence levels. To further highlight the performance differences, we report the Prediction Interval Coverage Probability (PICP) and the Average Width Deviation (AWD) under the same settings, as shown in Fig.~\ref{fig:PICP} and Fig.~\ref{fig:AWD}. The mean and standard deviation of PICP and AWD over the three runs are summarized in Table~\ref{tab:PICP_AWD}.

\begin{table}[pos=htbp]
\centering
\caption{PICP and AWD (\text{mean} $\pm$ std) under different confidence levels}
\label{tab:PICP_AWD}
\begin{minipage}{0.48\linewidth}
\centering
\text{(a) PICP}
\vspace{2mm}
\begin{tabular}{lccc}
\toprule
Model & 80\% & 90\% & 95\% \\
\midrule
ND 
& $0.686 \pm {\scriptsize 0.28}$ 
& $0.800 \pm {\scriptsize 0.24}$ 
& $0.871 \pm {\scriptsize 0.21}$ \\
TD 
& $0.636 \pm {\scriptsize 0.28}$ 
& $0.746 \pm {\scriptsize 0.25}$ 
& $0.827 \pm {\scriptsize 0.22}$ \\
HQ
& $0.815 \pm {\scriptsize 0.24}$ 
& $0.895 \pm {\scriptsize 0.19}$ 
& $0.940 \pm {\scriptsize 0.14}$ \\
\bottomrule
\end{tabular}
\end{minipage}
\hfill
\begin{minipage}{0.48\linewidth}
\centering
\text{(b) AWD}
\vspace{2mm}
\begin{tabular}{lccc}
\toprule
Model & 80\% & 90\% & 95\% \\
\midrule
ND 
& $0.434 \pm {\scriptsize 0.71}$ 
& $0.269 \pm {\scriptsize 0.56}$ 
& $0.172 \pm {\scriptsize 0.45}$ \\
TD 
& $0.508 \pm {\scriptsize 0.72}$ 
& $0.328 \pm {\scriptsize 0.59}$ 
& $0.219 \pm {\scriptsize 0.48}$ \\
HQ
& $0.226 \pm {\scriptsize 0.51}$ 
& $0.119 \pm {\scriptsize 0.38}$ 
& $0.071 \pm {\scriptsize 0.30}$ \\
\bottomrule
\end{tabular}
\end{minipage}
\end{table}

For example, at the 80\% confidence level, HQ achieves a PICP of 81.5\%, which is closest to the nominal confidence level compared with ND (68.6\%) and TD (63.6\%). Meanwhile, HQ obtains the lowest AWD of 0.226, corresponding to reductions of 47.9\% and 40.3\% compared with ND and TD, respectively.
The PICP results show that HQ yields coverage probabilities that are closest to the nominal confidence levels with minimal deviation, indicating better calibration and reliability. In contrast, ND and TD exhibit larger discrepancies from the nominal levels.
Moreover, AWD quantifies the deviation of targets that fall outside the prediction intervals. HQ consistently achieves the smallest AWD across all confidence levels, suggesting higher-quality intervals with fewer severe misses.

\subsection{Validation experiment}
\subsubsection{Confidence Calibration}
\begin{figure}[pos=htbp]
\centering
\begin{subfigure}{0.48\linewidth}
    \centering
    \includegraphics[width=\linewidth]{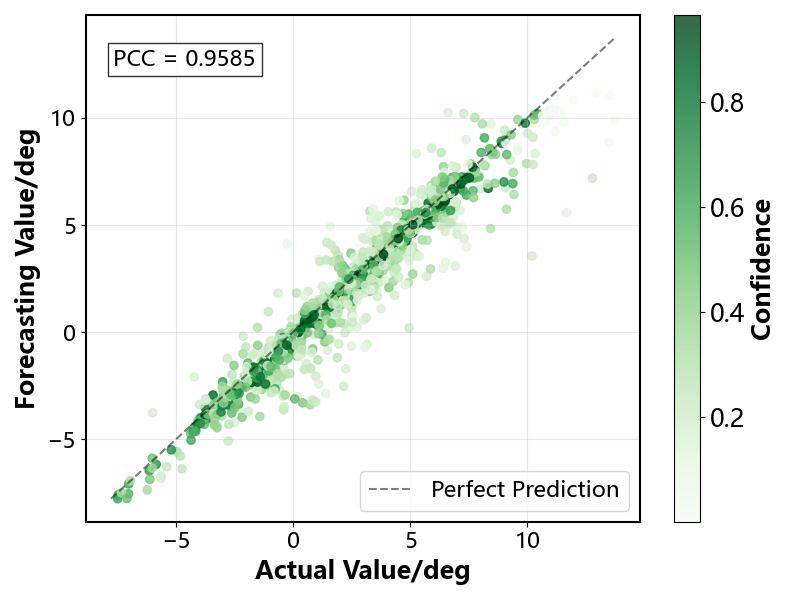}
    \caption{10-steps}
    \label{fig:parity-10steps}
\end{subfigure}
\hfill
\begin{subfigure}{0.49\linewidth}
    \centering
    \includegraphics[width=\linewidth]{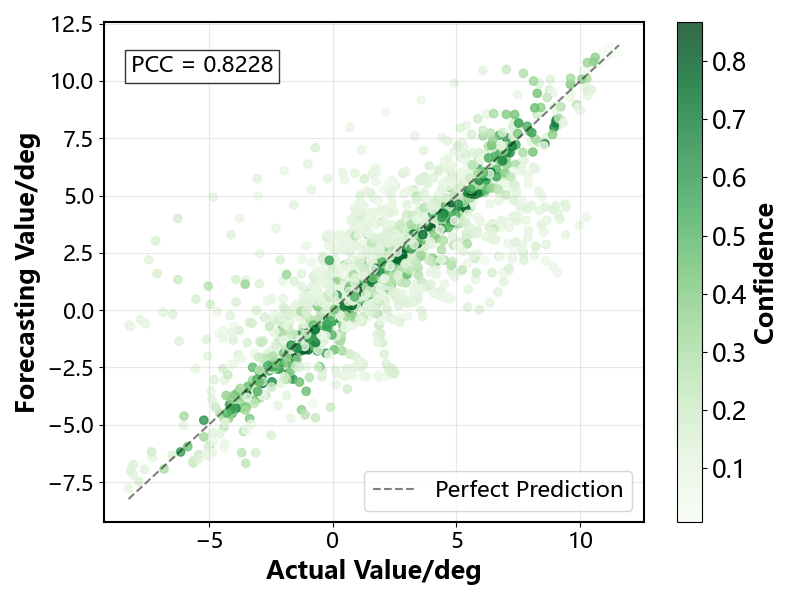}
    \caption{20-steps}
    \label{fig:parity-20steps}
\end{subfigure}
\caption{Parity plot (predicted vs. ground truth) with the identity line (y=x); point transparency indicates the predicted confidence (higher opacity denotes higher confidence)}
\label{fig:Parity plot}
\end{figure}

To further validate that the proposed confidence estimates are well calibrated and consistent with prediction accuracy, Fig.~\ref{fig:Parity plot} presents confidence-weighted parity plots for the 10-step and 20-step forecasts.

As shown in Fig.~\ref{fig:parity-10steps} and Fig.~\ref{fig:parity-20steps}, most scatter points are densely clustered around the dashed identity line ($y=x$), indicating that the predicted roll angles are in close agreement with the ground truth. This observation is further supported by the Pearson correlation coefficients (PCCs) of 0.9585 and 0.8228 for the 10-step and 20-step prediction, respectively, confirming a strong linear correlation and high overall prediction consistency under the proposed paradigm.

Moreover, point opacity encodes the estimated confidence level, where higher opacity (less transparency) indicates higher confidence. A clear trend can be observed: samples closer to the identity line generally appear more opaque, whereas those farther away tend to be more transparent. This spatial--confidence coupling suggests that the model assigns higher confidence to low-error predictions and lower confidence to high-error predictions, demonstrating a well-calibrated and interpretable confidence assessment mechanism for multi-step roll forecasting.

\begin{figure}[pos=htbp]
\centering
\begin{subfigure}{0.4\linewidth}
    \centering
    \includegraphics[width=\linewidth]{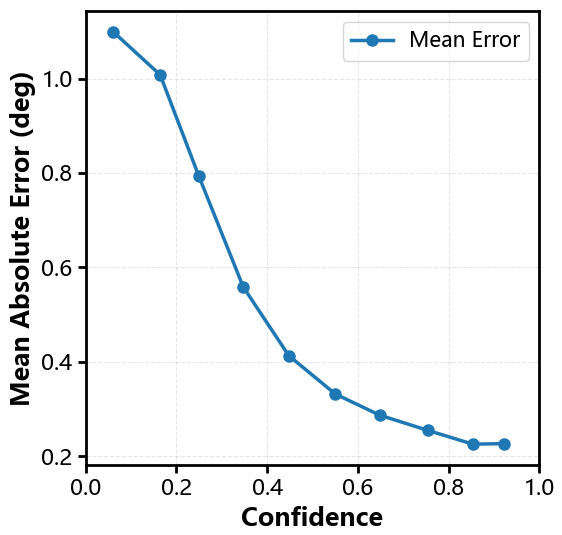}
    \caption{10-steps}
    \label{fig:Conf-error-10steps}
\end{subfigure}
\hfill
\begin{subfigure}{0.425\linewidth}
    \centering
    \includegraphics[width=\linewidth]{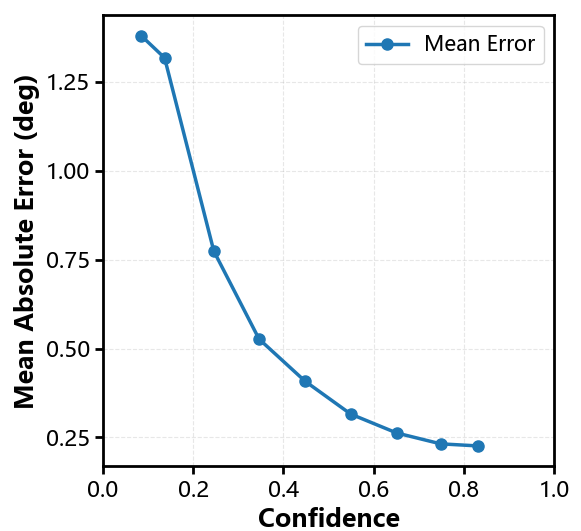}
    \caption{20-steps}
    \label{fig:Conf-error-20steps}
\end{subfigure}
\caption{The relationship of Error and Confidence}
\label{fig:Conf-error}
\end{figure}

To evaluate whether the estimated confidence is consistent with the actual forecasting performance, an error-confidence analysis was conducted, as shown in ref.Fig.~\ref{fig:Conf-error}. 
Specifically, the entire confidence interval $[0, 1]$ was uniformly partitioned into 10 sub-intervals with a bin step of 0.1. For each prediction, the absolute error was evaluated to the ground truth, and they were categorized into their corresponding confidence bins. Within each populated bin, the confidence scores and absolute errors were separately averaged, yielding mean confidence score and mean absolute error. These results were subsequently plotted to construct the final error-confidence calibration curve.
A monotonic decrease in prediction error with increasing confidence indicates that the confidence estimates effectively reflect the reliability of the forecasts. 

\subsubsection{Effectiveness of the consistency loss}

To evaluate the effectiveness of the proposed consistency loss and to verify whether the regression head and the classification head yield mutually consistent outputs, we conduct an ablation study with and without the consistency loss. In particular, we examine the alignment between the regression prediction (mapped to the corresponding discretized roll-bin index range) and the maximum-confidence region inferred from the classification head.

As shown in Table~\ref{tab:rmse of cons}, we evaluate the error variations before and after introducing the consistency loss using RMSE metrics. the introduction of the consistency loss leads to a slight increase in RMSE for the 20-5, 20-10, and 20-15 prediction horizons, with increases of 3.97\%, 3.53\%, and 0.49\%, respectively. This is because the consistency constraint encourages the alignment between regression and classification outputs, which improves the reliability and uncertainty estimation capability at the cost of a marginal increase in point prediction error.

\begin{figure}[pos=htbp]
\centering
\includegraphics[width=0.8\linewidth]{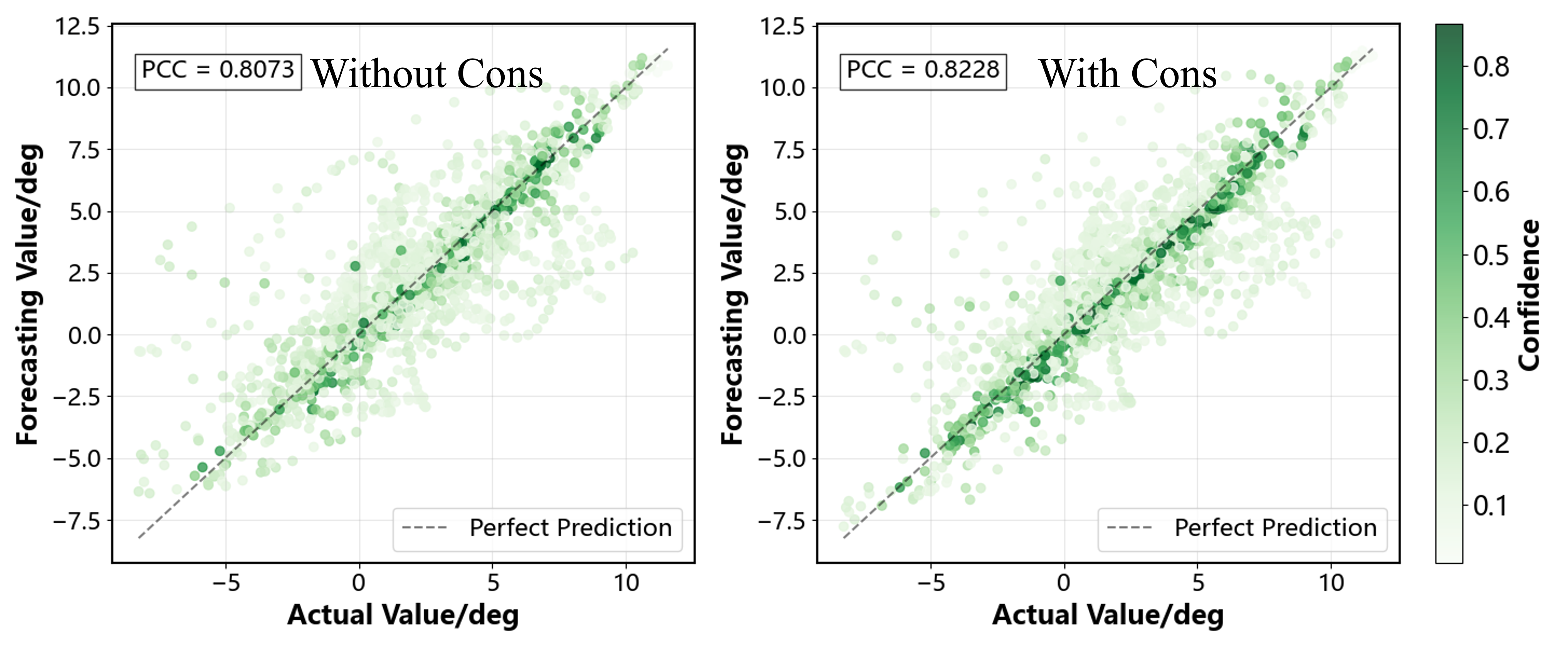}
\caption{Ablation results with and without the proposed consistency loss. }
\label{fig:Ablation of consistency loss}
\end{figure}

As shown in Fig.~\ref{fig:Ablation of consistency loss}, incorporating the consistency loss increases the proportion of high-confidence predictions concentrated near the identity line ($y=x$). This indicates that the consistency loss effectively improves the agreement between the regression outputs and the high-confidence bins predicted by the classification head, thereby enhancing the reliability and interpretability of the joint regression--classification framework.

\begin{table}[pos=htbp]
  \centering 
   \caption{Variation of RMSE with and without the consistency loss}
  \label{tab:rmse of cons}
  \resizebox{0.5\textwidth}{!}{ 
  \begin{tabular}{ccccc} 
    \toprule
     In/Output & 20-5 & 20-10 & 20-15 & 20-20  \\    
    \midrule
    Without Cons & 0.327 & 1.104 &1.831&2.214   \\
    With Cons& 0.340 & 1.143 &1.840&2.175  \\
    Rate(\%) & 3.97 & 3.53 & 0.49 &-1.76\\
    \bottomrule
  \end{tabular}
  }
\end{table}

\subsection{Significance test}
To verify whether the performance improvements of the proposed framework are statistically significant, Wilcoxon signed-rank test was conducted. RMSE and MAE were evaluated for point prediction performance, while PICP and AWD under different confidence levels (80\%, 90\%, and 95\%) were analyzed for uncertainty quantification performance.

As shown in Table~\ref{tab:normalization_significance}, AdaC performs significantly better than most normalization methods, with
$p$-values below 0.05 for both RMSE and MAE. The only exception is RevIN on MAE ($p=0.2529$), where the improvement is not
significant.
Table~\ref{tab:hq_significance} shows the significance results of HQ compared with ND and TD at different confidence levels. HQ
performs significantly better in both PICP and AWD at all confidence levels, with all $p$-values below 0.05.
These results suggest that the performance gains are not due to random changes during training, but come from the effectiveness
of the proposed framework.
\begin{table}[pos=htbp]
\centering
\caption{Significance analysis of AdaC against different normalization strategies.}
\label{tab:normalization_significance}
\renewcommand{\arraystretch}{1.15}
\begin{tabular}{cccccc}
\toprule
Model & \multicolumn{2}{c}{RMSE} & \multicolumn{2}{c}{MAE} \\
\cmidrule(lr){2-3} \cmidrule(lr){4-5}
& $p$-value & Significant & $p$-value & Significant \\
\midrule
No-standardization    
& 0.0006 & Yes & 0.0013 & Yes \\
Z-Score 
& 0.0016 & Yes & 0.0015 & Yes \\
RevIN   
& 0.0413 & Yes & 0.2529 & No  \\
Dish-ts 
& 0.0093 & Yes & 0.0083 & Yes \\
San     
& 0.0097 & Yes & 0.0103 & Yes \\
\bottomrule
\end{tabular}
\end{table}

\begin{table}[pos=htbp]
\centering
\caption{Significance analysis of HQ against ND and TD under different confidence levels.}
\label{tab:hq_significance}
\renewcommand{\arraystretch}{1.15}
\begin{tabular}{cccccc}
\toprule
Confidence level & Comparison & \multicolumn{2}{c}{PICP} & \multicolumn{2}{c}{AWD} \\
\cmidrule(lr){3-4} \cmidrule(lr){5-6}
& & $p$-value & Significant & $p$-value & Significant \\
\midrule
\multirow{2}{*}{80\%}
&  ND & $3.78\times10^{-21}$ & Yes & $1.34\times10^{-21}$ & Yes \\
&  TD & $1.81\times10^{-32}$ & Yes & $2.85\times10^{-31}$ & Yes \\
\midrule
\multirow{2}{*}{90\%}
&  ND & $1.17\times10^{-17}$ & Yes & $3.10\times10^{-17}$ & Yes \\
&  TD & $1.23\times10^{-30}$ & Yes & $2.25\times10^{-28}$ & Yes \\
\midrule
\multirow{2}{*}{95\%}
&  ND & $2.07\times10^{-14}$ & Yes & $4.91\times10^{-14}$ & Yes \\
&  TD & $3.35\times10^{-26}$ & Yes & $5.27\times10^{-24}$ & Yes \\
\bottomrule
\end{tabular}
\end{table}

\section{Conclusion and discussion}
\subsection{Conclusion}
To address the lack of explicit reliability quantification in USVs roll prediction, this study proposes a reliability-aware forecasting paradigm that augments a regression-based predictor with an adaptive feature processing stage and an auxiliary confidence-assessment branch. Specifically, the adaptive centralization strategy is employed to mitigate non-stationary mean drift in short-term roll sequences, thereby improving generalization under varying operating conditions, while the confidence-assessment head leverages discretized roll bins to provide step-wise confidence scores and prediction intervals at different nominal confidence levels. In this way, the proposed paradigm not only produces accurate roll forecasts but also delivers actionable reliability information for downstream USVs decision-making.

Experiments on real-world USVs datasets demonstrate that the proposed adaptive centralization significantly enhances generalization across operating conditions, and that the estimated confidence is consistent with the observed prediction errors. Moreover, the overall paradigm is model-agnostic and offers a certain degree of interpretability. 
\subsection{Limitations and future work}
Although the proposed paradigm achieves reliability-aware prediction and robust generalization in USV roll forecasting, several limitations must be noted:
First, the proposed model is primarily designed for short-term prediction, and its performance in long-term prediction scenarios may degrade in terms of both accuracy and confidence reliability, as longer prediction horizons naturally introduce greater uncertainty and cumulative error.
Second, due to current experimental limitations, the framework only utilizes historical roll motion sequences without incorporating external environmental factors, such as wind speed and wave height. Therefore, under rapidly varying sea conditions, relying solely on historical dynamics may limit the model’s ability to capture sudden roll peaks.

Several directions remain for future work. First, the coupling relationships between roll motion and multiple input variables (e.g., wind speed, wave height) can be considered to boost the model performance. Second, future research will explore integrating the proposed framework with mechanistic models to further improve interpretability and physical consistency.

\section*{Funding}
The financial support from the National Natural Science Foundation of China (No.62225308), and Shanghai Science and Technology Innovation Project (No.XTCX-KJB002-2025-2-03) is also acknowledged.

\section*{Data availability statement}
The data that support the findings of this study are openly available in
Open Science Framework at \url{https://osf.io/huazc/overview}.

\section*{Disclosure statement}
The authors report there are no competing interests to declare.

\bibliographystyle{cas-model2-names}
\bibliography{Ref}

\end{document}